\documentclass{article}
\usepackage[top=40truemm,bottom=40truemm,left=30truemm,right=30truemm]{geometry}  

\usepackage{titling}
\usepackage{authblk}
\usepackage{microtype}
\usepackage{graphicx}
\usepackage{subfigure}
\usepackage{booktabs} 

\usepackage{hyperref}

\usepackage[utf8]{inputenc} 
\usepackage[T1]{fontenc}    
\usepackage{url}            
\usepackage{amsfonts}       
\usepackage{nicefrac}       
\usepackage{amsmath,amsthm,amssymb}

\usepackage{natbib}
\usepackage{comment,color}
\usepackage{bm}
\usepackage{wrapfig}
\usepackage{adjustbox}
\usepackage{algorithm}
\usepackage{algorithmic}

\usepackage{lipsum}

\newcommand\blfootnote[1]{%
  \begingroup
  \renewcommand\thefootnote{}\footnote{#1}%
  \addtocounter{footnote}{-1}%
  \endgroup
}

\renewcommand{\H}{{\mathcal{H}}}

\newtheorem{proposition}{Proposition}

\newcommand{\tk}[1]{   {\color{magenta}{TK: #1}}  }
\newcommand{\mk}[1]{   {\color{blue}{MK: #1}}  }

\newcommand{\argmax}{\operatornamewithlimits{argmax}}

\begin{document}

\author[1,2,5]{Takafumi Kajihara* }
\author[3,2]{Motonobu Kanagawa* }
\author[1]{Yuuki Nakaguchi}
\author[1]{Kanishka Khandelwal}
\author[4,2]{Kenji Fukumizu}
\affil[1]{NEC Corporation}
\affil[2]{RIKEN AIP}
\affil[3]{University of T\"ubingen and Max Planck Institute for Intelligent Systems}
\affil[4]{The Institute of Statistical Mathematics}
\affil[5]{National Institute of Advanced Industrial Science and Technology}
\date{}                   
\setcounter{Maxaffil}{0}
\renewcommand\Affilfont{\itshape\small}

\title{Model Selection for Simulator-based Statistical Models:\\ A Kernel Approach}
\maketitle

\blfootnote{* Equal contribution. Correspondence to: Takafumi Kajihara <t-kajihara@ct.jp.nec.com>, Motonobu Kanagawa <motonobu.kanagawa@tuebingen.mpg.de>}




\begin{abstract}
We propose a novel approach to model selection for simulator-based statistical models.
The proposed approach defines a mixture of candidate models, and then iteratively updates the weight coefficients for those models as well as the parameters in each model simultaneously; this is done by recursively applying Bayes' rule, using the recently proposed kernel recursive ABC algorithm.
The practical advantage of the method is that it can be used even when a modeler lacks appropriate prior knowledge about the parameters in each model.
We demonstrate the effectiveness of the proposed approach with a number of experiments, including model selection for dynamical systems in ecology and epidemiology. 
\end{abstract}

\section{Introduction}
\label{sec:introduction}

Computer simulation is nowadays a ubiquitous tool in modern scientific and engineering problems to study time-evolving complex phenomena, such as climate modeling, social science, ecology and epidemiology \citep{Winsberg2010,Weisberg2012}.
In the language of statistics, computer simulation can be defined as sampling from a probabilistic model that describes the phenomena of interest.
One practically important issue regarding simulation is how to select an appropriate model, if there exist multiple candidate models. 
This problem of {\em model selection} is challenging in various ways, in the context of computer simulation.
From the viewpoint of statistics, a major obstacle comes from the incapability of using a likelihood function; this is because the conditional probability $P_m(y|\theta^m)$ of data $y$ given model parameters $\theta^m$ cannot be calculated in general, as the data-generating process is often very complicated (where we use $m = 1,\dots,K$ for an indicator of one of $K$ candidate models.)
For instance, this is the case where simulation involves numerically solving differential equations describing the dynamical system of interest; see Sec.~\ref{sec:experiments} for examples in ecology and epidemiology.



{\em Approximate  Bayesian Computation} (ABC) is a simulator-based approach to Bayesian inference, which is applicable as long as sampling from a probabilistic model is possible \cite{tavare1997inferring,pritchard1999population,beaumont2002approximate,SisFanBea18}.
ABC has been used in model selection for simulator-based statistical models (\citealt{grelaud2009abc, toni2009simulation,toni2009approximate, pudlo2015reliable}; \citealt[Chapter~6]{SisFanBea18});
see Supplements for a brief review. 
Being a Bayesian approach, these ABC-based methods select a model by essentially comparing the {\em marginal likelihood} (or the {\em evidence}) $\int P_m(y|\theta^m)  \pi_m(\theta^m)d\theta^m$ for each of the candidate models, provided a prior distribution $\pi_m(\theta^m)$ of the model parameters $\theta^m$.
Note that in the Bayesian approach, a ``model'' $m$ is given by the tuple $(P_m, \pi_m)$ including the prior $ \pi_m(\theta)$, i.e., the hierarchical model of first generating parameters $\theta^m \sim \pi_m(\theta)$ and then data $y \sim P_m(y|\theta^m)$; the marginal likelihood $\int P_m(y|\theta^m) d\pi_m(\theta^m)$ represents how likely the tuple $(P_m, \pi_m)$ is true.

The Bayesian approach, however, may not always be the best way in practice, if a modeler lacks appropriate prior knowledge about parameters in a model.
For instance, consider a situation where a modeler only has vague prior knowledge about a certain parameter $\theta^m$ that it should be positive and ``not very large,'' say it cannot be larger than $1000$, and suppose that she sets her prior $\pi_m(\theta^m)$ as the uniform distribution over $[0, 1000]$.
But why should it be a uniform distribution?
The choice of the uniform prior just reflects her lack of knowledge, and she might have chosen it because there was no better alternative.
Therefore, the form of the prior $\pi_m(\theta^m)$ in this case is not fully informative, and thus it may not be the best idea to define the ``model'' as the tuple $(P_m, \pi_m)$ and perform model selection by faithfully relying on the marginal likelihood $\int P_m(y|\theta^m) d\pi_m(\theta^m)$.

In this work, we rather focus on situations where a modeler is willing to define  $P_m(y|\theta^m)$ as a ``model''  and is interested in selecting one model from candidates and estimating its parameters.
For instance, this is the case where one is interested in predictions or forecasting of phenomena of interest; in such a case, a modeler is asked to select one model and its parameters that result in the best predictive performances. 
This is true in particular when simulations are computationally expensive, situations common when modeling complex phenomena (e.g.,~spatio-temporal modeling in climate science and multi-agent systems in social science). 
In such a case one can only afford to have one good model with tuned parameters.
On the other hand, a (fully) Bayesian approach requires taking averages over candidate models and parameters therein, which is practically not possible.

In this paper, we propose a novel approach to model selection for simulation-based statistical models that simultaneously estimates model parameters. 
We propose to define a mixture of candidate models and associate each model with a weight coefficient, which intuitively represents how that model is likely to be true; those weights and the parameters in each model are simultaneously and iteratively updated, by recursively applying Bayes' rule.
This idea is implemented with the recently proposed kernel recursive ABC algorithm \citep{pmlr-v80-kajihara18a}, a kernel-based method for point estimation with simulator-based statistical models.



This paper is organized as follows.
We first explain the conceptual idea of our approach in Sec.~\ref{sec:proposed-concept}, and then present a concrete algorithm based in Sec.~\ref{sec:proposed-concrete}.
We report a number of experimental results to empirically assess the proposed approach in Sec.~\ref{sec:experiments}, including model selection for dynamical systems in ecology and epidemiology. 
A review of related approaches, details of experiments and additional results are provided in the Supplementary Materials. 



\section{Proposed approach: the conceptual idea}\label{sec:proposed-concept}
In this section we present the conceptual idea of our approach to model selection, based on which we propose a concrete algorithm in Sec.~\ref{sec:proposed-concrete}.

Assume that there are $K \geq 2$ candidate models.
Each of the models $m \in \{1,\dots,K\}$ has a conditional distribution of observed data $y \in \mathcal{Y}$ given model parameters $\theta^m \in \Theta^m$, denoted by $P_m(y | \theta^m )$, where $\mathcal{Y}$ and $\Theta^m$ are measurable spaces.
We also assume that for each model $m$ there is a prior distribution $\pi_m(\theta^m)$ for the model parameters.
Suppose we are given observed data $y^* \in \mathcal{Y}$.

\subsection{Mixture of candidate models}
Our idea is to combine multiple models to form a mixture model.
The proposed method then estimates the mixing coefficients, which represent the  ``probabilities'' of the models being true, as well as the parameters in the models simultaneously. 

Let $\phi^1, \dots, \phi^K \geq 0$ be mixing coefficients such that $\sum_{m=1}^K \phi^m = 1$ and that each coefficient $\phi_m$ is associated with the model $P_m(y|\theta^m)$ of the same model index $m = 1,\dots,K$.
We introduce the notation $\Psi$ to write all the mixing coefficients and the associated model parameters:
$$
\Psi := \left( (\phi^m)_{m=1}^K , (\theta^m)_{m=1}^K \right) \in S_{K-1}  \times \Theta_1 \times \cdots \times \Theta_K,
$$
where $S_{K-1} := \{ (\phi^1, \dots, \phi^K) \in \mathbb{R}^K: \phi_1,\dots,\phi_K \geq 0,\ \sum_{m=1}^K \phi^m = 1.\}$ is the $K-1$ simplex.
That is, $\Psi$ is defined on the product space of $S_{K-1}$ and all the parameter spaces $\Phi_1,\dots,\Phi_K$.
We then define a mixture of the candidate models as
\begin{equation} \label{eq:mixture_model}
P(y | \Psi) := \sum_{m=1}^K \phi^m P_m(y | \theta^m).
\end{equation}
In other words, to generate $y$ from this model, we first sample a model index $m$ from the multinomial distribution with coefficients $\phi^1,\dots,\phi^K$, and then sample $y$ from $P_m(y|\theta^m)$.
For simplicity, we assume each model $P_m(y | \theta^m)$ has a density function $p_m(y | \Psi)$, and we denote by $p(y | \Psi)$ the resulting density function of \eqref{eq:mixture_model}.
Note that in the setting of this paper $p(y^* | \Psi)$ cannot be evaluated, but in the sequel assume that we can obtain the resulting posterior densities in order to explain the idea.

\subsection{Recursive Bayes update for the mixture}
\label{sec:recursive_bayes}

Our idea is to recursively apply Bayes' rule to the same observed data $y^*$ to obtain a sequence of posterior distributions of $\Psi$ {\em on the product space} $S_{K-1} \times \Theta_1 \times \cdots \times \Theta_K$.
That is, we set the posterior distribution derived at one iteration as the prior for the next iteration.

For the first recursion $N = 1$, we define a prior distribution $\pi^1(\Psi)$ on $S_{K-1} \times \Theta_1 \times \cdots \times \Theta_K$ as 
\begin{equation} \label{eq:initial-prior}
\pi^1 (\Psi) := \mathrm{Dir}( (\phi^m)_{m=1}^K |\alpha) \prod_{m=1}^K \pi_m(\theta^m), 
\end{equation}
where $\mathrm{Dir}( (\phi^m)_{m=1}^K |\alpha)$ denotes the Dirichlet distribution with concentration parameter $\alpha > 0$, and $\pi_m(\theta^m)$ is the prior of the parameter in each model.
Then the posterior $p^{1}(\Psi | y^*)$ at the first recursion is given by Bayes' formula
$$
p^{1}(\Psi | y^*) \propto p(y^* | \Psi) \pi^1(\Psi).
$$
For $N \geq 2$, assume now that the posterior $p^{N-1}(\Psi | y^*)$ at the $(N-1)$-th recursion has already been obtained.
Define a prior $\pi^N(\Psi)$ at the $N$-th recursion by this previous posterior:
\begin{equation} \label{eq:prior-recursion}
\pi^N(\Psi) := p^{N-1}(\Psi | y^*).
\end{equation}
Then the posterior distribution at the $N$-th recursion is again given by Bayes' rule 
\begin{equation} \label{eq:posterior-recursion}
p^N( \Psi | y^* ) \propto p(y^* | \Psi) \pi^N( \Psi ).
\end{equation}
Our motivation of considering the recursive Bayes updates may be understood by rewriting \eqref{eq:posterior-recursion} as
$$
p^N( \Psi | y^* ) \propto \left( p(y^* | \Psi) \right)^N  \pi^1( \Psi ).
$$
This shows that, as $N \to \infty$, the powered likelihood term $ \left( p(y^* | \Psi) \right)^N$ gradually dominates the posterior $p^N( \Psi | y^* )$, and the effects of the prior $\pi^1( \Psi )$ gradually diminish.
In fact, \citet[Corollary to Lemma A.2 in p.1624]{LelNadSch10} implies that the power posterior $p^N( \Psi | y^* )$ degenerates to the Dirac distribution at the maximum likelihood point $\psi_{\rm max}:=\argmax_{\Psi}p(y^* | \Psi)$ as $N \to \infty$, if it exists.
Therefore, one may expect that the recursive Bayes updates provide a way of obtaining $\Psi = \left( (\phi^m)_{m=1}^K , (\theta^m)_{m=1}^K \right)$ such that the largest coefficient in $(\phi^m)_{m=1}^K$, say $\phi^{m*}$, indicates the ground-truth model and the associated parameters $\theta^{m^*}$ are those of maximum likelihood.
This is an intuitive idea why we introduce the recursive Bayes updates; mathematically there is a subtle issue in this argument, which we will discuss in Sec.~\ref{sec:proposed-concrete}.
We nevertheless provide one theoretical result below that indicates how the recursive Bayes updates take model complexities take into account. 



\begin{proposition} \label{prop:posterior-expansion}
Let $p^N ( (\phi^m)_{m=1}^K | y^* )$ and $\pi^N( (\phi^m)_{m=1}^K )$ be the marginal densities of $(\phi^m)_{m=1}^K$ induced from the posterior  \eqref{eq:posterior-recursion} and the prior \eqref{eq:prior-recursion} (or \eqref{eq:initial-prior} if $N = 1$), respectively.
Then for $N \geq 2$, we have
\begin{eqnarray}
 p^N ( (\phi^m)_{m=1}^K  | y^* ) =  C_N^{-1} \pi^N( (\phi^m)_{m=1}^K )  \times
  \sum_{m=1}^K  \phi^m \int  p_m( y^* | \theta^m ) \pi^N(\theta^m | (\phi^i)_{i=1}^K ) d\theta^m, \label{eq:posterior-expnasion}
\end{eqnarray}
where $C_N := \int p(y^* | \Psi) \pi^N( \Psi ) d\Psi$ is a normalization constant, and $\pi^N(\theta^m | (\phi^i)_{i=1}^K )$ is the conditional density of $\theta^m$ given $(\phi^i)_{i=1}^K$ induced from the prior \eqref{eq:prior-recursion} if $N \geq 2$, and $\pi^N(\theta^m | (\phi^i)_{i=1}^K ) = \pi_m(\theta^m)$ if $N = 1$.
\end{proposition}

In Proposition \ref{prop:posterior-expansion}, the identity \eqref{eq:posterior-expnasion} describes how the posterior density of $(\phi^m)_{m=1}^K$ changes from $\pi^N( (\phi^m)_{m=1}^K ) = p^{N-1}( (\phi^m)_{m=1}^K | y^* )$ to $p^N( (\phi^m)_{m=1}^K | y^* )$ as a result of one application of Bayes' rule.
The key insight is that the integral 
\begin{equation} \label{eq:marginal-likelihood-recursion}
\int p_m( y^* | \theta^m ) \pi^N(\theta^m | (\phi^i)_{i=1}^K ) d\theta^m
\end{equation}
can be regarded as the {\em marginal likelihood or model evidence} for the model $m$, for which the prior of the model parameter $\theta^m$ is updated to $\pi^N(\theta^m | (\phi^i)_{i=1}^K ) = p^{N-1}(\theta^m | (\phi^i)_{i=1}^K, y^* )$.
In fact, for $N = 1$ this interpretation is more evident, since in this case the integral is $\int p_m( y^* | \theta^m ) \pi_m(\theta^m) d\theta^m$, the marginal likelihood of the model $m$ with the original prior $\pi_m(\theta^m)$.

To provide an intuition about how the update of the posterior of $(\phi_m)_{m=1}^K$ occurs, let us  assume that the updated prior in \eqref{eq:marginal-likelihood-recursion} is independent of the coefficients $(\phi^i)_{i=1}^K$, i.e., $\pi^N(\theta^m | (\phi^i)_{i=1}^K ) = \pi_m^N(\theta^m)$, where $\pi_m^N(\theta^m)$ is the marginal density of $\theta^m$ induced from \eqref{eq:prior-recursion}. 
Then the integral \eqref{eq:marginal-likelihood-recursion} is $\int p_m( y^* | \theta^m ) \pi_m^N(\theta^m ) d\theta^m$, and the update \eqref{eq:posterior-expnasion} becomes
\begin{eqnarray*}
 p^N ( (\phi^m)_{m=1}^K  | y^* ) =  C_N^{-1} \pi^N( (\phi^m)_{m=1}^K ) \times 
  \sum_{m=1}^K  \phi^m \int  p_m( y^* | \theta^m ) \pi_m^N(\theta^m ) d\theta^m.
\end{eqnarray*}
Assume $K=2$, and $\int  p_1( y^* | \theta^1 ) \pi_1^N(\theta^1 ) d\theta^1 > \int  p_2( y^* | \theta^2 ) \pi^N(\theta^2 ) d\theta^2$, i.e., the marginal likelihood of the model 1 is larger than that of the model 2.
Let $( \tilde{\phi}^1, \tilde{\phi}^2 ) \in S_1$.
If $\tilde{\phi}^1 > \phi^1$ (and thus $\tilde{\phi}^2 < \phi^2$), then we have 
$ \sum_{m=1}^2  \tilde{\phi}^m \int  p_m( y^* | \theta^m ) \pi_m^N(\theta^m ) d\theta^m >  \sum_{m=1}^2  \phi^m \int  p_m( y^* | \theta^m ) \pi_m^N(\theta^m ) d\theta^m$.
Recalling the notation $\pi^N( (\phi^1, \phi^2) ) = p^{N-1} ( (\phi^1, \phi^2) | y^* )$, this implies that
$p^N( (\tilde{\phi}^1, \tilde{\phi}^2) | y^* ) / p^{N-1}( (\tilde{\phi}^1, \tilde{\phi}^2) | y^* ) > p^N( (\phi^1, \phi^2) | y^* ) / p^{N-1}( (\phi^1, \phi^2) | y^* )$.
In other words, if the marginal likelihood of the model 1 is larger than that of the model 2 at the current recursion, then the density $p^{N}( (\tilde{\phi}^1, \tilde{\phi}^2) | y^* )$ of a point in the simplex $(\tilde{\phi}^1, \tilde{\phi}^2) \in S_1$ with larger coefficient $\tilde{\phi}_1$ for the model 1 becomes relatively higher from the previous one $ p^{N-1}( (\tilde{\phi}^1, \tilde{\phi}^2) | y^* )$.
In this way, the mass of the posterior $p^N ( (\phi^m)_{m=1}^K  | y^* )$ may move towards the model with larger marginal likelihoods, as $N$ increases.

Note however that, in the above discussion, the assumption $\pi^N(\theta^m | (\phi^i)_{i=1}^K ) = \pi_m^N(\theta^m)$ is valid for the initial update (i.e., $N = 1$), but in general this may not hold. Therefore the argument is correct for the initial update, but for the case $N > 1$ its validity is unclear.
We leave theoretical analysis of the assumption $\pi^N(\theta^m | (\phi^i)_{i=1}^K ) = \pi_m^N(\theta^m)$ (or the properties of $\pi^N(\theta^m | (\phi^i)_{i=1}^K )$ for $N \geq 2$) for future research.

\section{Proposed approach: the concrete algorithm}
\label{sec:proposed-concrete}

To develop a concrete algorithm based on the above idea, we employ the kernel recursive ABC (KR-ABC) \citep{pmlr-v80-kajihara18a}, a kernel-based method for point estimation with intractable likelihood.
This method performs point estimation by recursively applying Bayes' rule in a reproducing kernel Hilbert space (RKHS).
More specifically, it iterates the two steps: 1) posterior computation with kernel ABC, and 2) deterministic sampling from the posterior via kernel herding.
There are two main reasons why we adopt this approach: i) To the best of our knowledge, this is the only existing method for recursive Bayes updates that can be used with intractable likelihood; ii) This method has a special property of being robust to misspecification of a prior distribution, which is the desideratum in our context.

\textbf{Kernel mean embedding.}
Before describing our method, we need to explain {\em kernel mean embedding of distributions} \citep{smola2007hilbert}, a framework for representing probability distributions in an RKHS.
For details of the following concepts, see \citet{MuaFukSriSch17}.
Let $\mathcal{X}$ be a measurable space, $\mathcal{P}$ be the set of all probability distributions on $\mathcal{X}$, $k: \mathcal{X} \times \mathcal{X} \to \mathbb{R}$ be a measurable, bounded, positive definite kernel and $\mathcal{H}$ be its RKHS.
In this framework, each distribution $P \in \mathcal{P}$ is represented as a Bochner integral
\begin{equation} \label{eq:kernel_mean}
\mu_P := \int k(\cdot, x)dP(x) \in \mathcal{H},
\end{equation}
which is referred to as the {\em kernel mean} (or the {\em mean embedding}) of $P$. 
The mapping $P \in \mathcal{P} \to \mu_P \in \mathcal{H}$ is one-to-one, if $\mathcal{H}$ is large enough; in such a case $k$ is called {\em characteristic}, an example being the Gaussian kernel $k(x,x') = \exp( - \| x - x' \|^2 / \gamma^2)$ on $\mathcal{X} \subset \mathbb{R}^d$ \citep{FukGreSunSch08,SriGreFukSchetal10}.
Thus if $k$ is characteristic, kernel mean $\mu_P$ is uniquely associated with $P$ and thus maintains all of its information.
This fact justifies estimation of $\mu_P$ in place of $P$, and estimation of $\mu_P$ is often easier than that of $P$ itself, thanks to the reproducing property of the RKHS $\mathcal{H}$.
For instance, given an i.i.d.~sample $X_1,\dots,X_n$ from $P$, $\mu_P$ is estimated by $\hat{\mu}_P := \frac{1}{n} \sum_{i=1}^n k(\cdot,X_i)$ with convergence rate $\mathbb{E}[\| \hat{\mu}_P - \mu_P \|_\mathcal{H}] = O(n^{-1/2})$, a rate independent of the dimensionality of $\mathcal{X}$ \citep{TohSriMua17}.

\begin{algorithm}[t]
   \caption{Model selection via KR-ABC}
   \label{al:model-select-kr-abc}
\begin{algorithmic}[1]
   \STATE {\bfseries Input:} 
    Observed data $y^* \in \mathcal{Y}$, 
    models $( P_m(y | \theta^m))_{m=1}^K$, priors $(\pi_m(\theta^m))_{m=1}^K$.
 \STATE {\bfseries Setting:} 
   Dirichlet concentration parameter $\alpha > 0$,
   number $N_{\rm iter}$ of iterations, 
   number $n$ of simulations per iteration,
   kernel $k$ on $\mathcal{X} = S_{K-1} \times \Theta^1 \times \cdots \times \Theta^K$,
   kernel $k_\mathcal{Y}$ on $\mathcal{Y}$, regularization constant $\delta > 0$.
   \FOR{\texttt{$N = 1,..., N_{\rm iter}$}} 
   \IF{$N=1$}{ \label{algo:if-prior-initial}
    	\FOR{\texttt{$i = 1,..., n$}}
        \STATE {\em Sampling}: $(\phi_{N,i}^m)_{m=1}^K \sim \mathrm{Dir} ( (\phi_m)_{m=1}^K |\alpha)$. \label{algo:dirichlet}
    	\STATE {\em Sampling}: $\theta_{N,i}^m \sim \pi_m (\theta^m)$, $m=1,\dots,K$. \label{algo:prior-initial}
        \ENDFOR
        }    
    \ENDIF \label{algo:ifend-prior-initial}
    \FOR{\texttt{$i = 1,..., n$}} \label{algo:for-simulate-observations}
    \STATE {\em Sampling}: $y_{N,i} \sim \sum_{m=1}^K \phi_{N,i}^m P_m(y |\theta_{N,i}^m)$. \label{algo:simulate-observations}
    \ENDFOR  \label{algo:forend-simulate-observations} 
    \STATE {\em Kernel Bayes}: compute ${\bm w} = (w_1,\dots,w_n)^T \in \mathbb{R}^n$ by Eq.(\ref{eq:KRR_weight}) using $G := (k_\mathcal{Y}(y_{N,i}, y_{N,j}))_{i,j=1}^n \in \mathbb{R}^{n \times n}$ and ${\bm k}(y^*) := (k_\mathcal{Y}(y_{N,i}, y^*))_{i=1}^n \in \mathbb{R}^n$. \label{algo:weight}
    \STATE {\em Deterministic sampling}: generate $(\Psi_{N+1,i})_{i=1}^n = ( (\phi_{N+1,i}^m)_{m=1}^K , ( \theta_{N+1,i}^m )_{m=1}^K ) )_{i=1}^n$ by kernel herding Eqs.(\ref{eq:herding-first}) (\ref{eq:herding_alg}) using kernel $k$, $(w_i)_{i=1}^n$ and $(\Psi_{N,i})_{i=1}^n = ( (\phi_{N,i}^m)_{m=1}^K , ( \theta_{N,i}^m )_{m=1}^K ) )$. \label{algo:herding}
\ENDFOR
   \STATE {\bfseries Output:} Final states
   $(\Psi_{N_{\rm iter}+1,i})_{i=1}^n = ( (\phi_{N_{\rm iter}+1,i}^m)_{m=1}^K , ( \theta_{N_{\rm iter}+1,i}^m )_{m=1}^K ) )_{i=1}^n \subset \mathcal{X}$. \label{algo:output}
\end{algorithmic}
\end{algorithm}

We are now ready to explain the proposed algorithm, which is summarized in Algorithm \ref{al:model-select-kr-abc}.
For the ease of explanation, we may call $\mathcal{X} := S_{K-1} \times \Theta^1 \times \cdots \times \Theta^K$ the {\em state space}, and $\Psi = ( (\phi^m)_{m=1}^K, (\theta^m)_{m=1}^K ) \in \mathcal{X}$ a {\em state}.
In the algorithm, $n \in \mathbb{N}$ denotes the number of simulations in each recursion, and $N_{\rm iter} \in \mathbb{N}$ the total number of recursions, which are specified by a user.

\textbf{Spaces and kernels.}
In our algorithm there are two kernels.
One kernel, denote by $k_\mathcal{Y}$, is defined 
on the space $\mathcal{Y}$ of summary statistics of observed data.
The other kernel, denoted by $k$, is defined on the product space $\mathcal{X} = S_{K-1} \times \Theta^1 \times \cdots \times \Theta^K$.
In particular we define $k$ as a product kernel composed of a kernel $k_{S_{k-1}}$ on $S_{k-1}$ and kernels $k_m$ respectively defined on $\Theta^m$.
More precisely, for $\Psi = ( (\phi^m)_{m=1}^K, (\theta^m)_{m=1}^K ), \tilde{\Psi} = ( (\tilde{\phi}^m)_{m=1}^K, (\tilde{\theta}^m)_{m=1}^K ) \in S_{K-1} \times \Theta^1 \times \cdots \times \Theta^m$, the kernel is defined as $k( \Psi, \tilde{\Psi} ) = k_{S_{k-1}}(  (\phi^m)_{m=1}^K, (\tilde{\phi}^m)_{m=1}^K ) \prod_{m=1}^K k_m( \theta^m, \tilde{\theta}^m )$.
In the sequel let $\mathcal{H}$ be the RKHS of this kernel $k$. 
In our experiments in Sec.~\ref{sec:experiments}, we focus on the case where each of $\mathcal{Y}$ and $\Theta^1, \dots, \Theta^K$ is a subset of a Euclidean space, and we define $k_\mathcal{Y}$ and $k_1,\dots,k_K$ as Gaussian kernels.
While there are kernels designed for simplex such as information-diffusion kernels \citep{LafLeb05},  we also use a Gaussian kernel as $k_{S_{k-1}}$; in our preliminary experiments we found the use of the information diffusion kernels somehow did not lead to better performance.

\textbf{Initial prior sampling (Lines 5-8):}
For the first recursion $N = 1$, the algorithm generates initial states $(\Psi_{N,i})_{i=1}^n = ( (\phi_{N,i}^m)_{m=1}^K, (\theta_{N,i}^m)_{m=1}^K )$ by randomly sampling from the prior $\pi^1(\Psi)$ in \eqref{eq:initial-prior};  $(\Psi_{N,i})_{i=1}^n$ provides an empirical approximation of $\pi^1(\Psi)$.

\textbf{1.~Bayes' rule via Kernel ABC (Lines 10-13):} \label{sec:kernelABC}
The aim of this step is to estimate the kernel mean 
\begin{equation} \label{eq:posterior-mean}
\mu_{p_N(\Psi | y^*)} := \int k(\cdot,\Psi) p_N(\Psi | y^*)d\Psi
\end{equation}
of the posterior $p^N(\Psi | y^*)$ \eqref{eq:posterior-recursion}.
To this end, we first obtain $n$ pairs of state and simulated data $( \Psi_{N,i}, y_{N,i} )_{i=1}^n$ by simulating data $y_{N,i}$ from state $\Psi_{N,i}$ using the mixture model \eqref{eq:mixture_model} for $i=1,\dots,n$.
(If $N \geq 2$, assume that states $(\Psi_{N,i})_{i=1}^n$ have already been generated at the Step 2 of deterministic posterior sampling in the $(N-1)$-th recursion, explained below.)
Since $(\Psi_{N,i})_{i=1}^n$ approximately follow the prior $\pi^N(\Psi)$ \eqref{eq:prior-recursion}, $( \Psi_{N,i}, y_{N,i} )_{i=1}^n$ approximately follow the joint distribution $p(y | \Psi) \pi^N( \Psi )$.
Note that the posterior \eqref{eq:posterior-recursion} is the conditional distribution of $\Psi$ given $y^*$ induced from this joint distribution.
Therefore by estimating the conditional distribution of $\Psi$ given $y^*$ using $( \Psi_{N,i}, y_{N,i} )_{i=1}^n$, one can estimate the posterior \eqref{eq:posterior-recursion}; this is the idea of ABC.

Kernel ABC \citep{nakagome2013kernel} instead estimates the kernel mean of this conditional distribution using $(\Psi_{N,i}, y_{N,i} )_{i=1}^n$, and this results in the following estimator of the posterior kernel mean \eqref{eq:posterior-mean}: 
\begin{eqnarray}
&\hat{\mu}_{p_N(\Psi | y^*)} := \sum^{n}_{i=1}w_i k(\cdot , \Psi_{N,i})  \in \mathcal{H},& \label{eq:cond_kmean} \\
&{\bm w} := (w_1,\dots,w_n)^T := (G + n \delta I)^{-1} {\bm k}(y^*),& \label{eq:KRR_weight}
\end{eqnarray}
where ${\bm k}(y^*) := (k_\mathcal{Y}(y_{N,1},y^*), \dots, k_\mathcal{Y}(y_{N,n}, y^*))^T \in \mathbb{R}^n$, $G := (k_\mathcal{Y}(y_{N,i},y_{N,j}))_{i,j=1}^n \in \mathbb{R}^{n \times n}$, $\delta > 0$ is a regularization constant, and $I \in \mathbb{R}^{n \times n}$ is the identity. 
The consistency and convergence rates of \eqref{eq:cond_kmean} as $n \to \infty$, which require $\delta \to 0$ with appropriate speed, have been studied extensively in the literature; see \citet{GruLevBalPatetal12,fukumizu2013kernel,MuaFukSriSch17}.

\textbf{2.~Deterministic posterior sampling via kernel herding (Line 14):} \label{sec:kernelherding}
In this step, the algorithm generates new states $(\Psi_{N+1,i})_{i=1}^n$ from \eqref{eq:cond_kmean}, so that $(\Psi_{N+1,i})_{i=1}^n$ approximately follow the posterior \eqref{eq:posterior-recursion}; these new states will be used in the next recursion to simulate observed data (Line 11).
One approach for this purpose is {\em kernel herding} \citep{chen2010super}, a class of deterministic sampling method that is a greedy variant of Quasi Monte Carlo \citep{DicKuoSlo13}.
\citet{pmlr-v80-kajihara18a} showed that the use of kernel herding in the recursive Bayes updates leads to the robustness against a misspecified prior.

The concrete procedure, which is greedy, is as follows.
The first state is given by 
\begin{equation} \label{eq:herding-first}
\Psi_{N+1,1} := \argmax_{\Psi \in \mathcal{X}} \sum_{i=1}^{n} w_i k(\Psi, \Psi_{N,i}),  
\end{equation}
where $w_1,\dots,w_n$ are given by \eqref{eq:KRR_weight}.
Assume now that $\Psi_{N+1,1}, \Psi_{N+1,2}, \dots, \Psi_{N+1,t}$ for $1 \leq t < n$ have already been generated.
Then the next state $\Psi_{N+1,t+1}$ is 
\begin{eqnarray} 
\Psi_{N+1,t+1} 
&:=& \argmax_{\Psi \in \mathcal{X}} \sum_{i=1}^{n} w_i k(\Psi, \Psi_{N,i})  \label{eq:herding_alg} \\
&& -  \dfrac {1}{t+1}\sum^t_{i=1}k(\Psi , \Psi_{N+1,i}). \nonumber
\end{eqnarray}
In this way, one can obtain $n$ states $(\Psi_{N+1,i})_{i=1}^n$.
It can be shown that if $k$ is shift-invariant (e.g., Gaussian), this procedure \eqref{eq:herding-first} \eqref{eq:herding_alg} is equivalent to the greedy minimization of the RKHS distance 
$
\left\| \hat{\mu}_{p_N(\Psi | y^*)} - \frac{1}{t} \sum_{i=1}^t k(\cdot,\Psi_{N+1,i})  \right\|_{\mathcal{H}}
$
between $\hat{\mu}_{p_N(\Psi | y^*)}$ and the empirical average $\frac{1}{t} \sum_{i=1}^t k(\cdot,\Psi_{N+1,i})$ for $t = 1,\dots,n$ \citep{chen2010super}.
In terms of this distance the convergence rate $O(n^{-1/2})$ is guaranteed for $\frac{1}{n} \sum_{i=1}^n k(\cdot,\Psi_{N+1,i})$ \citep{bach2012equivalence}, which may hold even when the optimization problem \eqref{eq:herding-first} \eqref{eq:herding_alg} are solved approximately \citep{LacLinBac15,KanNisGreFuk16}.
Since i) $\mu_{p_N(\Psi | y^*)}$ contains the all information of $p_N(\Psi | y^*)$, ii) $\hat{\mu}_{p_N(\Psi | y^*)}$ is a consistent estimator of $\mu_{p_N(\Psi | y^*)}$, and iii) $\frac{1}{n} \sum_{i=1}^n k(\cdot,\Psi_{N+1,i})$ is a convergent approximation of $\hat{\mu}_{p_N(\Psi | y^*)}$, we can say that the states $(\Psi_{N+1,i})_{i=1}^n$ approximately follow the posterior $p_N(\Psi | y^*)$; this is in the sense that the empirical average $\frac{1}{n}\sum_{i=1}^n f(\Psi_{N+1,i})$ converges to the expectation $\int f(\Psi) p_N(\Psi | y^*) d\Psi$ as $n \to \infty$ for all $f \in \mathcal{H}$ or for any $f$ that can be approximated well by functions in $\mathcal{H}$ \citep{KanSriFuk16}.

\textbf{Output (Line 16):}
After $N_{\rm iter}$ recursions, the algorithm yields the states $(\Psi_{N_{\rm iter}+1,i})_{i=1}^n = ( (\phi_{N_{\rm iter}+1,i}^m)_{m=1}^K , ( \theta_{N_{\rm iter}+1,i}^m )_{m=1}^K ) )_{i=1}^n \subset \mathcal{X}$, which are an approximate sample of the posterior $p^{N_{\rm iter}}(\Psi | y^*)$.
While there can be various ways to exploit these states, in this paper we propose to use the first state $\Psi_{N_{\rm iter}+1,1} = ((\phi_{N_{\rm iter}+1,1}^m)_{m=1}^K , ( \theta_{N_{\rm iter}+1,1}^m )_{m=1}^K )$ for the purpose of model selection and parameter estimation.
This is motivated from \citet[Prop.~2]{pmlr-v80-kajihara18a}, which shows that under certain assumptions the first state of kernel herding \eqref{eq:herding-first} converges to the mode of the posterior $p^{N_{\rm iter}}(\Psi | y^*)$ as $N_{\rm iter}$ increases.
Therefore $\Psi_{N_{\rm iter}+1,1}$ can be regarded as an approximation of the mode of $p^{N_{\rm iter}}(\Psi | y^*)$.

For instance, in our experiments we select the model $m$ with the largest coefficient $\phi_{N_{\rm iter}+1,1}^m$ as a ``true'' model, and the associated $\theta_{N_{\rm iter}+1,i}^m$ as an estimate of the true parameter.
We note that the magnitudes of the coefficients $(\phi_{N_{\rm iter}+1,1}^m)_{m=1}^K$ are informative: For instance, if one of the coefficients is close to $1$ and the rest is nearly to $0$, then we may interpret the algorithm being confident that the ground-truth is the model associated with that largest coefficient; if all the coefficients are close to $1/K$, the algorithm's bet is neutral.

\textbf{Discussion.}
While the proposed method is an application of KR-ABC \citep{pmlr-v80-kajihara18a} to the mixture model \eqref{eq:mixture_model}, the principle we rely on is different from that in \citet{pmlr-v80-kajihara18a}.
In \citet{pmlr-v80-kajihara18a}, the goal is not model selection, but parameter estimation for a {\em given} parametric model with intractable likelihood.
The principle they rely on is \citet[Corollary to Lemma A.2 in p.1624]{LelNadSch10}, which claims that, if the likelihood function has a {\em unique} global maximum and satisfies other regularity conditions, then the recursive Bayes updates result in posteriors converging to the Dirac distribution at the maximum likelihood parameter.
While this assumption may be reasonable in the context of \citet{pmlr-v80-kajihara18a}, it may not hold in our case: For the mixture model \eqref{eq:mixture_model}, there is in general no unique global maximum.
Therefore, we cannot use the result of \citet{LelNadSch10} to analyze the convergence behavior of the posterior \eqref{eq:posterior-recursion} as $N$ goes to infinity.

While currently we do not have a complete characterization of the convergence property of the posterior \eqref{eq:posterior-recursion}, we attribute the model selection mechanism of our method to   Prop.~\ref{prop:posterior-expansion}:
As argued in Sec.~\ref{sec:recursive_bayes}, this preliminary result implies that the recursive Bayes updates for the mixture model \eqref{eq:mixture_model} may automatically take the marginal likelihoods of the candidate models into account.
We leave further theoretical analysis for future work, and empirically investigate the properties of  the proposed method by experiments in Sec.~\ref{sec:experiments}.

\section{Experiments}\label{sec:experiments}
We report results of experiments conducted for empirically assessing the proposed approach.
We first explain the setting common to all the experiments in Sec.~\ref{sec:exp-setting}.
We then report results on model selection for polynomial regression models in Sec.~\ref{sec:polynomial}, followed by results on dynamical systems in ecology (Sec.~\ref{sec:predator-prey}) and epidemiology (Sec.~\ref{sec:epidemics}).


\subsection{Experimental settings} \label{sec:exp-setting}

\textbf{Competing methods.}
We made comparisons with the following ABC-based methods: {\em ABC for Model Choice} (ABC-MC) \citep{grelaud2009abc}, {\em ABC Sequential Monte Carlo SMC} (ABC-SMC) \citep{toni2009approximate} and {\em ABC Random Forests} (ABC-RF) \citep{pudlo2015reliable}; see the Supplements for a review of these methods.


\textbf{Hyper-parameters.}
For each method, we determined its hyper-parameters by adopting a cross-validation-like approach described in \citet[Sec.~4]{Park2015}, unless otherwise stated; see the Supplements for details.
We set the Dirichlet concentration prior to be $0.01$ in the proposed method for all the experiments in the main body.

\textbf{Parameter estimation.}
For comparison, we also performed parameter estimation with the competing methods based on selected models; see the Supplements for details. 

\textbf{Evaluation measure.}
We performed experiments by varying the total number of simulations; the proposed method and ABC-SMC used 100 simulations for each iteration in all the experiments.
To evaluate each method, we computed the average number of mistakes made in model selection, where the average is taken over 30 independent trials of an experiment; we refer to this as the {\em model error}.
To further evaluate the quality in parameter estimation, we calculated the {\em data error} defined as follows:
for each of 30 trials, using the estimated parameters in a selected model we simulated pseudo data and measure the Euclidean distance to the observed data; then the data error is the average over these Euclidean distances over 30 trials. 


\subsection{Polynomial regression models} \label{sec:polynomial}
\label{sec:polynomials-misspecified}

We performed simple experiments using polynomial regression models.
We first investigated the effects of the Dirichlet concentration parameter to the behavior of the mixing coefficients.
The results suggest that the performance of the proposed method is stable w.r.t.~the choice of the concentration parameter; see the Supplements for details.

Here we investigate the robustness of the proposed approach against misspecification of prior distributions of model parameters. 
This is motivated from  \citet[Sec.~3]{pmlr-v80-kajihara18a}, who argue the use of kernel herding leads to an auto-correction mechanism that makes the method robust against misspecification of priors.

Let $x_1,\dots,x_{25} \in [0,5]$ be equally-spaced points such that $-1 = x_1 < x_2 < \cdots < x_{25} = 5$.
We consider two models: a 3rd polynomial model $y_i = \sum_{\ell=0}^3 a_\ell x_i + 3\varepsilon_i$ and a 4th order model $y_i = \sum_{\ell = 0}^{4} b_\ell x_i + 3\varepsilon_i$ $(i = 1,\dots,25)$, where $\varepsilon_1,\dots,\varepsilon_{25}$ are independent standard Gaussian noises and $(a_\ell)_{\ell=0}^4, (b_\ell)_{\ell=0}^{10} \subset \mathbb{R}$ are the parameters in the models.
We generated observed data $y^* := (y^*_1,\dots,y^*_{25}) \in \mathbb{R}^{25}$ from either of the two models, with parameters $a_0 = \cdots = a_3 = 40$ or $b_0 =  \cdots = b_4 = 40$.
We consider two cases where the prior distributions are (i) appropriate or (ii) misspecified.
Specifically, for each of parameters $(a_\ell)_{\ell=0}^{3}$ and $(b_\ell)_{\ell=1}^{4}$, we set a prior distribution as the uniform distribution over $[30,50]$ for (i) the appropriate case, and the uniform distribution over $[0,30]$ for (ii) the misspecified case.

\begin{figure}[t]
\begin{center}
\includegraphics[width=0.8\linewidth]{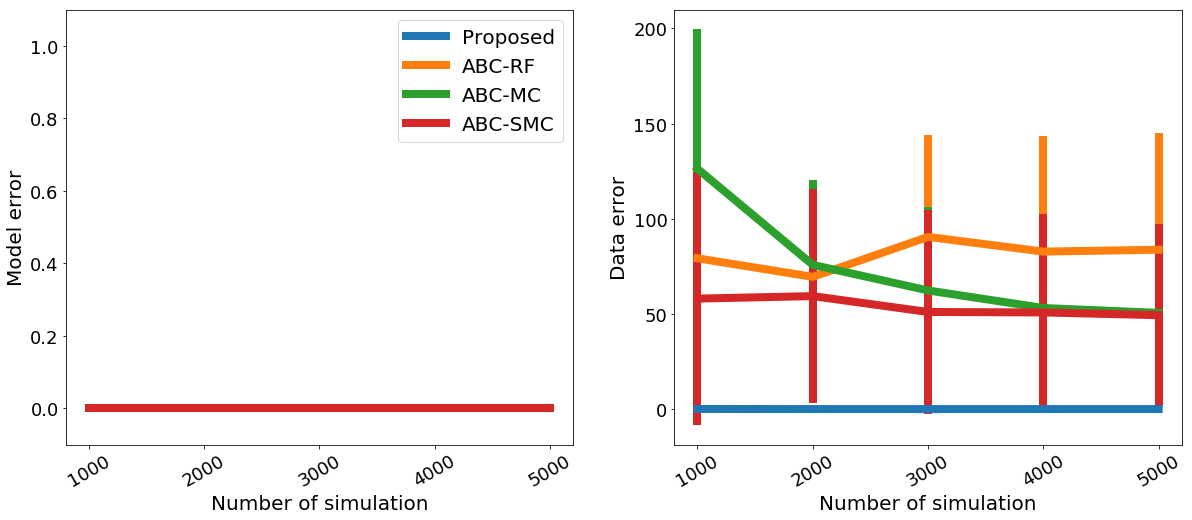}
\end{center}
\vspace{-3mm}
\caption{Model errors (left) and data errors (right) for varying total numbers of simulations in the appropriate case.
The ground truth is the 3rd order model. (Sec.~\ref{sec:polynomials-misspecified})}
\label{fig:poly_normal_1}
\vspace{-2mm}
\end{figure}

\begin{figure}
\begin{center}
\includegraphics[width=0.8\linewidth]{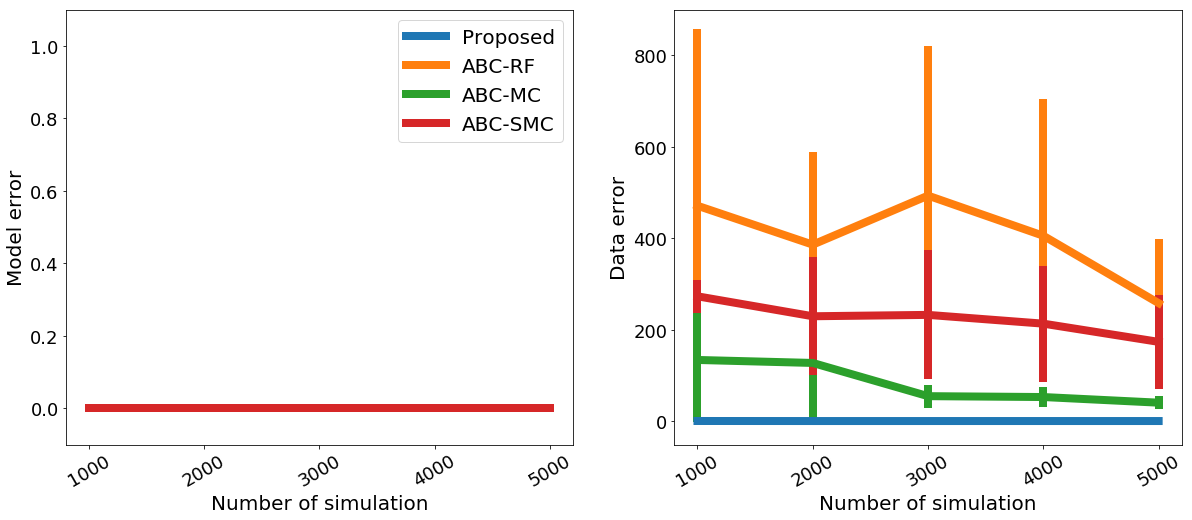}
\end{center}
\vspace{-3mm}
\caption{Model errors (left) and data errors (right) for varying total numbers of simulations in the appropriate case.
The ground truth is the 4th order model. (Sec.~\ref{sec:polynomials-misspecified})}
\label{fig:poly_normal_2}
\vspace{-2mm}
 \end{figure}

\begin{figure}[t]
 \begin{center}
\includegraphics[width=0.8\linewidth]{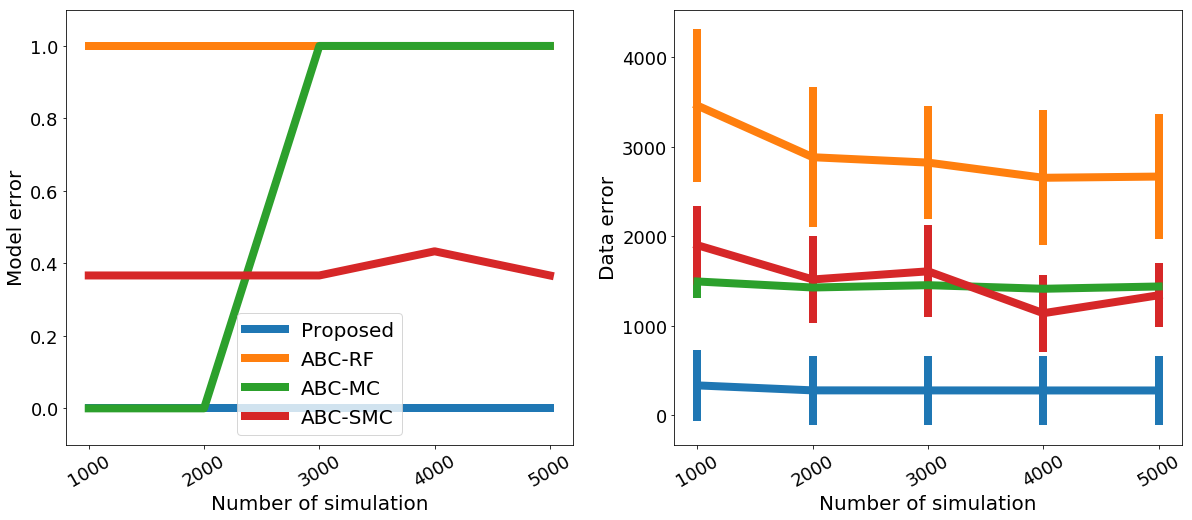}
\end{center}
\vspace{-3mm}
\caption{Model errors (left) and data errors (right) for varying total numbers of simulations in the misspecified case.
The ground truth is the 3rd order model. (Sec.~\ref{sec:polynomials-misspecified})} 
  \label{fig:poly_mis1}
  \vspace{-2mm}
\end{figure}

\begin{figure}[t]
\begin{center}
\includegraphics[width=0.8\linewidth]{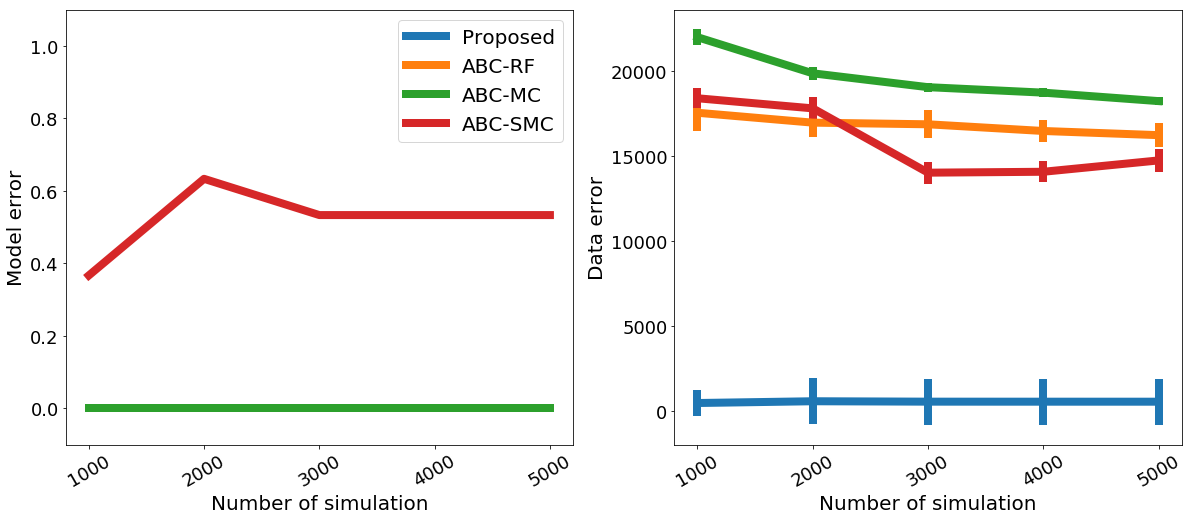} 
\end{center}
\vspace{-3mm}
\caption{Model errors (left) and data errors (right) for varying total numbers of simulations in the misspecifed case.
The ground truth is the 4th order model. Note that in the left figure, the proposed method, ABC-RF and ABC-MC made no mistakes and thus the lines are overlapping. (Sec.~\ref{sec:polynomials-misspecified})}
\label{fig:poly_mis2}
\vspace{-4mm}
 \end{figure}

Figures \ref{fig:poly_normal_1} and \ref{fig:poly_normal_2} show the results for (i) the appropriate case, while Figures \ref{fig:poly_mis1} and \ref{fig:poly_mis2} show those for (ii) the misspecified case. 
For the appropriate case, all the methods correctly identified the true model. 
On the other hand, for misspecified case, only the proposed method was able to correctly identify the true model, for both cases where the ground truth was the 3rd order or 4th order. 
The propose method also outperformed the other methods in terms of data errors for all the experiments. 
This is the same for errors in extrapolation; see the Supplements.


 %

\subsection{Predator-prey models in ecology}
\label{sec:predator-prey}

We conducted experiments on dynamical systems for predator-prey models studied in ecology; this is a typical example of intractable likelihood models. 
 Assume that there are two species, one being predators and the other being prey, and let $x(t) \geq 0$ and $y(t) \geq 0$ respectively represent the populations of prey and predators at time $t \in [0,T]$, where we set $T = 20$.
 In this section and Sec.~\ref{sec:epidemics}, we use ``dot'' to denote derivatives w.r.t.~$t$. (e.g., $\dot{x}(t)$ is the derivative of $x(t)$ w.r.t. $t$.)
 We consider the following two models defined by ordinal differential equations (ODE).

\textbf{Model 1}: The most fundamental model is the Lotka-Volterra equations  \citep{volterra1926variations,lotka1926elements}:
 \begin{eqnarray*}
\dot{x}(t) &=& a_1 x(t) - a_2 x(t) y(t), \\
 \dot{y}(t) &=&  - a_3 y(t) + a_4 x(t)y(t),
 \end{eqnarray*}
 where $a_1, a_2, a_3, a_4 \geq 0$ are the parameters of the model describing how the two specifies interact.

\textbf{Model 2}:
 The Lotka-Volterra model assumes that there are no intra-species interactions, but this may not be the case in reality: If the population of one species becomes too large, then there can be competition within that specifies for limited resources. 
Another model \citep{bazykin1998nonlinear} taking into account such intra-species interactions is given by
 \begin{eqnarray*}
  \dot{x}(t) &=& b_1 x(t) - b_2 x(t) y(t) - b_5 x^2(t),\\
  \dot{y}(t) &=& - b_3 y(t) + b_4 x(t) y(t) - b_6 y^2(t)
 \end{eqnarray*}
 where $b_1, b_2, b_3, b_4 \geq 0$ are the  parameters for inter-specifies interactions and  $b_5, b_6 \geq 0$ are parameters representing intra-species competition rates.


For both models, we defined the initial values as $x(0) = 10$ and $ y(0) = 5$.
When generating observed data, we used as ground-truth either Model 1 with $a_1 = 1$, $a_2 = 0.1$, $a_3 = 1.5$ and $a_4 = 0.75$ or Model 2 with $b_1 = 1$, $b_2 = 0.1$, $b_3 = 1.5$, $b_4 = 0.75$, $b_5 = 0.01$ and $b_6 = 0.01$. 
We set the prior distribution of each parameter in $(a_i)_{i=1}^4$ and $(b_i)_{i=1}^6$ as the uniform distribution over $[0,2]$. 

To generate data from the ODEs, we used the SciPy ODE solver (scipy.integrate.odeint) with step size 1 and all other parameters being the default values; this results in $T \times 2 = 40$ dimensional summary statistics of observations.
To make the problem more realistic, we added an independent standard Gaussian noise to each dimension of summary statistics, both for observed and simulated data.
Trajectories from the two models are described in the Supplements.



\begin{figure}[t]
  \begin{center}
\includegraphics[width=0.8\linewidth]{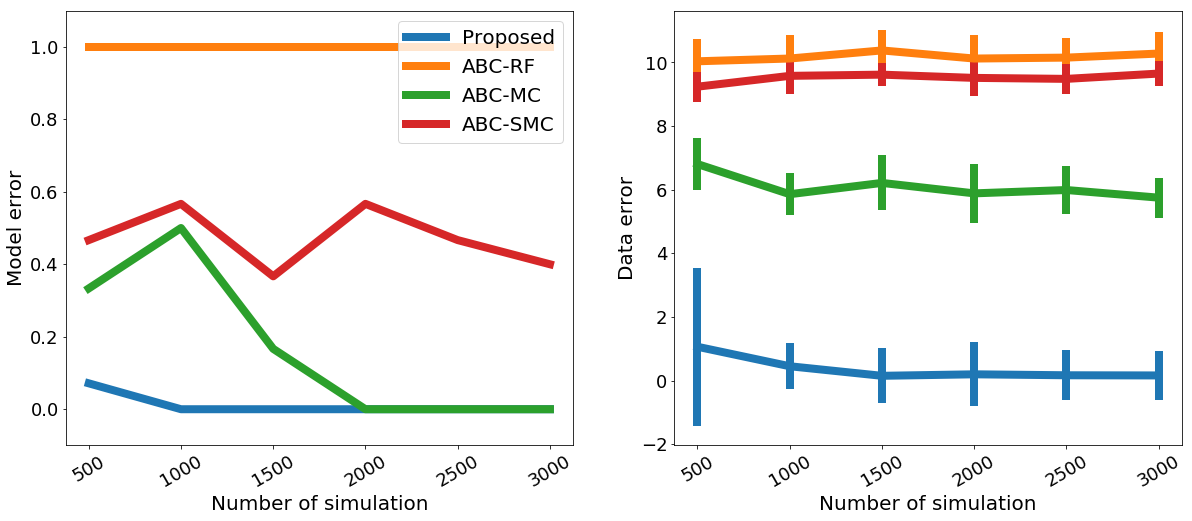}
  \end{center}
    \vspace{-3mm}
    \caption{Model errors (left) and data errors (right) for varying total numbers of simulations.
    The ground-truth is Model 2. (Sec.~\ref{sec:predator-prey})}
      \label{fig:predator-prey-complex-truth-normal}
      \vspace{-2mm}
 \end{figure}

The results are shown in Figures \ref{fig:predator-prey-simpler-truth-normal} and \ref{fig:predator-prey-complex-truth-normal}. 
Only the proposed method was able to reliably estimate the ground-truth model in both cases. 
Also in terms of data errors (and errors in extrapolation or prediction; see the Supplements), the proposed method significantly outperformed the competitors.  
We also conducted another experiment in a more difficult setting, and the proposed method performed the best; see the Supplements.


  \begin{figure}[t]
  \begin{center}
\includegraphics[width=0.8\linewidth]{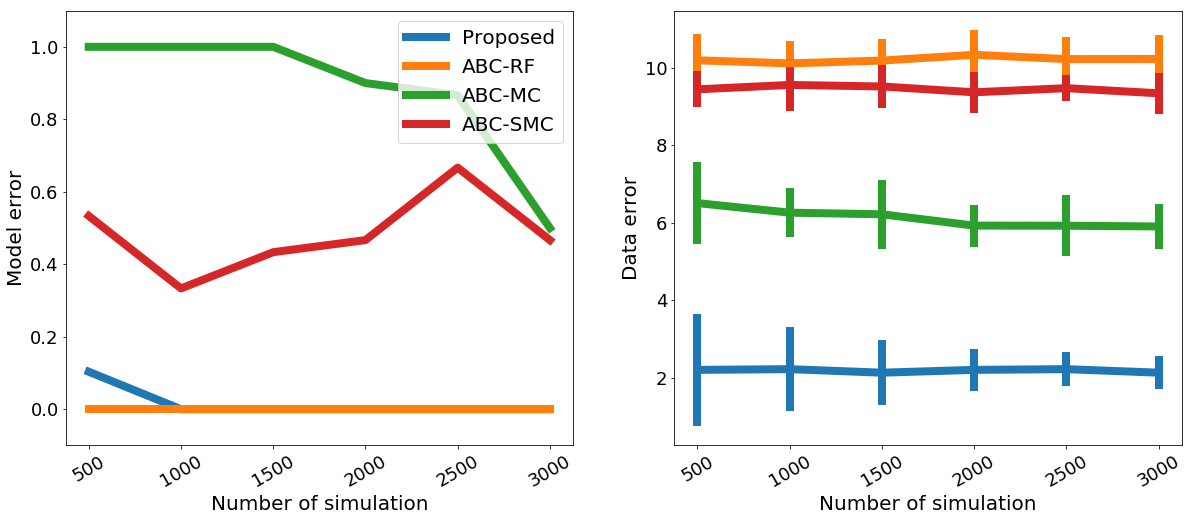}
  \end{center}
   \vspace{-3mm}
    \caption{Model errors (left) and data errors (right) for varying total numbers of simulations.
    The ground-truth is Model 1. 
    (Sec.~\ref{sec:predator-prey})}
      \label{fig:predator-prey-simpler-truth-normal}
      \vspace{-2mm}
 \end{figure}

\subsection{Dynamical systems for epidemics} \label{sec:epidemics} 

\begin{figure}[t]
  \begin{center}
\includegraphics[width=0.8\linewidth]{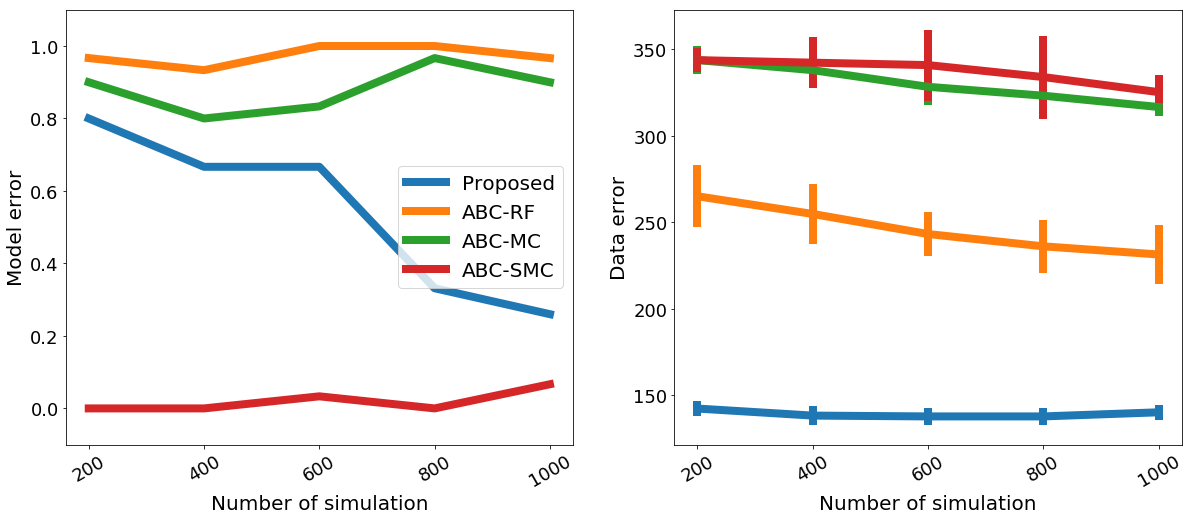}
  \end{center}
  \vspace{-3mm}
    \caption{Model errors (left) and data errors (right) for varying total numbers of simulations. 
    The ground-truth is Model 1. (Sec.~\ref{sec:epidemics})}
    \label{fig:sir_1}
 \end{figure}

\begin{figure}[t]
  \begin{center}
\includegraphics[width=0.8\linewidth]{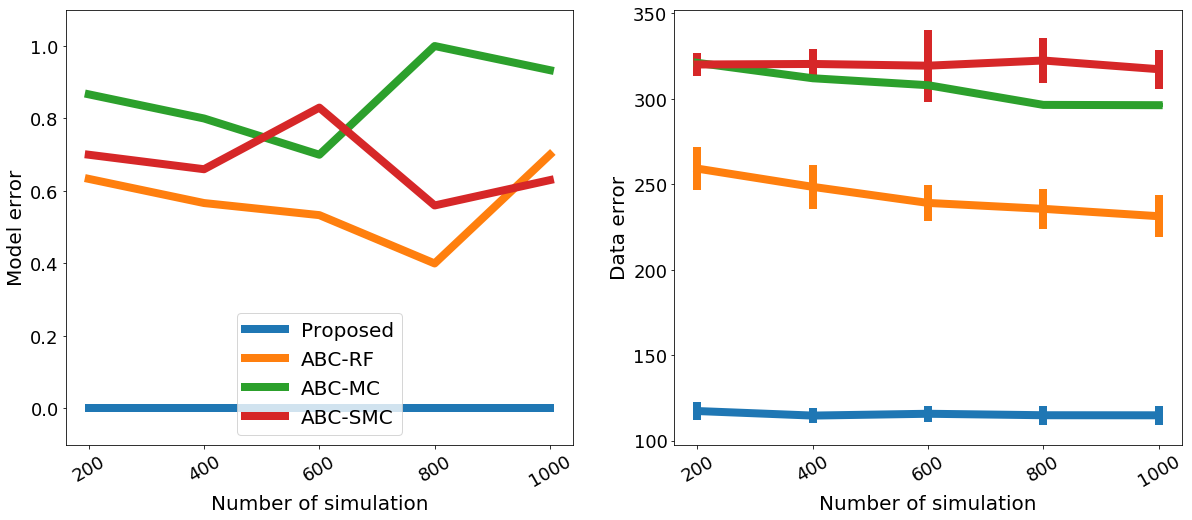}
  \end{center}
  \vspace{-3mm}
    \caption{Model errors (left) and data errors (right) for varying total numbers of simulations. 
    The ground-truth is Model 2. (Sec.~\ref{sec:epidemics})}
    \label{fig:sir_2}
 \end{figure}

 \begin{figure}[t]
  \begin{center}
\includegraphics[width=0.8\linewidth]{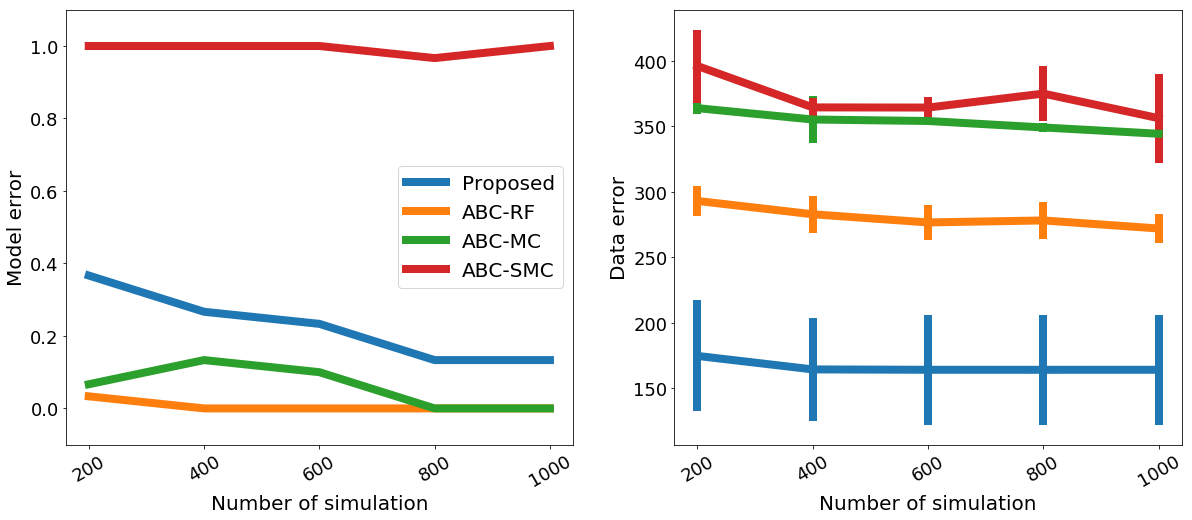}
  \end{center}
  \vspace{-3mm}
    \caption{Model errors (left) and data errors (right) for varying total numbers of simulations. The ground-truth is Model 3. (Sec.~\ref{sec:epidemics}) }
    \label{fig:sir_3}
 \end{figure}

In this experiment, we consider dynamical systems describing the spread of infectious diseases in a population of individuals. 
The most fundamental one is the SIR model \citep{anderson1992infectious}, which divides the population into susceptible (S), infected (I) and recovered (R) individuals.
Let $S(t)$, $I(t)$, and $R(t)$ be the numbers of individuals in these respective groups at time $t \in [0,T]$, where we set $T = 70$.
We consider the following three models defined by ODEs, following \citet[Sec.~3.3]{toni2009simulation}.
Below, the parameters $\alpha, \gamma, d, \nu \geq 0$ common for all the models respectively represent the rates for birth, infection, death, and recovery of individuals. 

\textbf{Model 1}: The first model considers a situation in which a newly infected individual (I) can immediately start infecting susceptible individuals (S):  
 \begin{eqnarray*}
 \dot{S}(t) &=& \alpha - \gamma S(t)I(t) - dS(t), \\
 \dot{I}(t) &=&  \gamma S(t)I(t) -\nu I(t) -dI(t), \\
 \dot{R}(t) &=&  \nu I(t) - dR(t).
 \end{eqnarray*}
 \vspace{-5mm}

\textbf{Model 2}: 
This model takes into account the situation where a newly infected individual goes into an unobserved Latent (L) state that cannot infect other susceptible individuals.
Thus the model includes a parameter $\delta \geq 0$ that represents the rate of transition from Latent to Infective state:
 \begin{eqnarray*}
 \dot{S}(t) &=& \alpha - \gamma S(t)I(t) - dS(t), \\
 \dot{L}(t) &=& \gamma S(t)I(t) - \delta L(t) - dL(t), \\
 \dot{I}(t) &=&  \delta L(t) -\nu I(t) -dI(t), \\
 \dot{R}(t) &=&  \nu I(t) - dR(t).
 \end{eqnarray*}
\textbf{Model 3}:
The third model allows for a recovered patient to be susceptible once again at a rate $e \geq 0$: 
 \begin{eqnarray*}
 \dot{S}(t) &=& \alpha - \gamma S(t)I(t) - dS(t) + eR(t), \\
 \dot{I}(t) &=&  \gamma S(t)I(t) -\nu I(t) -dI(t), \\
 \dot{R}(t) &=&  \nu I(t) - (d+e)R(t).
 \end{eqnarray*}
 
We set the initial values as $S(0) = 20$, $I(0) = 50$ and $R(0) = 0$ for all the models, and also $L(0) = 0$ for Model 2. 
When generating observed data, we set the parameters in the models as $\alpha = 0.5$, $\gamma = 0.001$, $d = 0.01$ and $\nu = 0.02$; also $\delta = 0.1$ (Model 2) and $e =0.1$ (Model 3). 
For each of parameters we set a prior distribution as the uniform distribution over $[0,1]$.

As in Sec.~\ref{sec:predator-prey}, we used the SciPy ODE solver with step size 1 with the default parameters to generate data from the ODEs, and this results in $T \times 3 = 210$ dimensional summary statistics of observations.
We added an independent standard Gaussian noise to each dimension of summary statistics, both for observed and simulated data.
Observed data from the three models are described in the Supplements.

The results are shown in Figures \ref{fig:sir_1}, \ref{fig:sir_2} and \ref{fig:sir_3}. 
In terms of model errors, the competitive methods performed well for some cases but completely failed for other cases.
For instance, ABC-SMC selected the correct model when the ground-truth was Model 1, but completely failed when the ground-truth was Model 3.
On the other hand, the proposed performed reasonably well for all the cases.
In terms of data errors (and extrapolation or prediction errors; see the Supplements), the proposed method significantly outperformed the other methods.
Additional experimental results in a more difficult setting are reported in the Supplements.


 

 


\section{Conclusions and future directions} \label{sec:conclusions}
For the problem of model selection for simulator-based statistical models, we proposed an approach based on a mixture of candidate models and recursive Bayesian updates, implemented using the kernel recursive ABC algorithm.
Empirical results suggest the validity of the proposed approach.

An obvious topic for future research is theoretical analysis: statistical convergence properties of the proposed algorithm should be revealed mathematically.
This also applies to the kernel recursive ABC algorithm itself, for which convergence guarantees have not been established. 
Another important direction is to investigate applications to large scale problems. 
This includes for instance multi-agent systems and spatio-temporal modeling involving partial differential equations; these have been used extensively in computer simulation in real-world problems, such as social science and climate modeling.
We hope that our work serves as one important step towards solving technical challenges arising in these problems.

\subsection*{Acknowledgements}
We thank Keisuke Yamazaki for fruitful discussions.
MK acknowledges support by the European Research Council (StG Action 757275 PANAMA). KF has been supported in part by JSPS KAKENHI 26280009.

\bibliographystyle{apalike}
\bibliography{kr_abc}

\newpage
\appendix
\begin{center}
    \bf{{\huge Supplementary Materials}}
\end{center}

\section{Proof of Proposition 1}

\begin{proof}
Below $\pi^N( (\theta^m)_{m=1}^K | (\phi^m)_{m=1}^K )$ denotes the conditional density of $(\theta^m)_{m=1}^K$ given $(\phi^m)_{m=1}^K$ that is induced from $\pi^N( (\phi^m)_{m=1}^K, (\theta^m)_{m=1}^K)$.  
\begin{eqnarray*}
 && p^N ( (\phi^m)_{m=1}^K  | y^* )  \\
 &=&  \int p^N ( (\phi^m)_{m=1}^K, (\theta^m)_{m=1}^K  | y^* ) d\theta^1 \cdots d\theta^K \\
&=& \int C_N^{-1}  p( y^* | (\phi^m)_{m=1}^K, (\theta^m)_{m=1}^K  )   \pi^N( (\phi^m)_{m=1}^K, (\theta^m)_{m=1}^K) d\theta^1 \cdots d\theta^K \\
&=& \int C_N^{-1} \left( \sum_{\ell=1}^K \phi^\ell p_\ell(y^* | \theta^\ell)  \right)   \pi^N( (\phi^m)_{m=1}^K)  \pi^N( (\theta^m)_{m=1}^K | (\phi^m)_{m=1}^K ) d\theta^1 \cdots d\theta^K \\
&=& C_N^{-1}  \pi^N( (\phi^m)_{m=1}^K)  \sum_{\ell=1}^K  \phi^\ell \int  p_\ell(y^* | \theta^\ell)   \pi^N(  (\theta^m)_{m=1}^K | (\phi^m)_{m=1}^K ) d\theta^1 \cdots d\theta^K \\
&=&  C_N^{-1} \sum_{\ell=1}^K  \phi^\ell \int  p_\ell(y^* | \theta^\ell) \pi^N( \theta^\ell | (\phi^m)_{m=1}^K ) d\theta^\ell.
\end{eqnarray*}
This proves the claim for the case $N \geq 2$.
For $N = 1$ the claim follows from the definition of $\pi^1( (\phi^m)_{m=1}^K, (\theta^m)_{m=1}^K )$ in (6) of the main text. 
\end{proof}

\section{A review of existing methods based on ABC} 
We provide a brief review of existing approaches to which we compare our method in the experiments.
To this end, for simplicity we restrict ourselves to the case where there are two candidate models, each of which consists of a conditional distribution $P_m(y|\theta^m)$ on $\mathcal{Y}$ given $\theta^m \in \Theta^m$ and a prior distribution $\pi_m(\theta^m)$ on $\Theta^m$, where $\mathcal{Y}$ and $\Theta^m$ are measurable spaces and $m = 1,2$. 
Assume that $P_m(y|\theta^m)$ has a density function $p_m(y|\theta^m)$.
Let $y^* \in \mathcal{Y}$ be observed data.

\paragraph{ABC for Model Choice (ABC-MC)} 
\label{sec:abcmc}
We first review the approach by \citet{grelaud2009abc} called {\em ABC for Model Choice (ABC-MC)}.
A standard Bayesian approach to model selection is to compute the Bayes factor \citep{kass1995bayes,robert2007bayesian} defined as the ratio between the marginal likelihoods of the two models:
$$
B_{1,2} := \frac{\int p_1(y^*|\theta^1) d\pi_1(\theta^1)}{\int p_2(y^*|\theta^2) d\pi_2(\theta^2)} = \frac{ \mathrm{Pr}(1|y^*) \mathrm{Pr}(2)}{ \mathrm{Pr}(2|y^*) \mathrm{Pr}(1) },
$$
where $\mathrm{Pr}(m|y^*)$ denotes the posterior probability of model $m$ being true, given observed data $y^*$, and $\mathrm{Pr}(m)$ is a prior probability of model $m$ being true; the second identity follows from Bayes' rule.
In ABC-MC, the Bayes factor is estimated using this second expression, estimating the two posterior probabilities $\mathrm{Pr}(m|y^*)$ by ABC.

Based on ABC, $\mathrm{Pr}(m|y^*)$ can be estimated in the following way. 
For $t = 1,\dots, n \in \mathbb{N}$, sample model index $m_t \sim \mathrm{Pr}(m)$ from the prior, sample model parameters $\theta_t^m \sim \pi_m(\theta^m)$, and then sample pseudo-data $y_t \sim P_m(y|\theta^m)$; if $\| y_t - y^*\| < \varepsilon$ for a prespecified threshold $\varepsilon > 0$ and a distance $\| \cdot \|$ defined on $\mathcal{Y}$, accept $y_t$; otherwise discard $y_t$.
Denoting by $(\tilde{m}_t)_{t=1}^T$ the set of the model indices associate with accepted data, where $T \leq n$ is the number of acceptances, the posterior probability $\mathrm{Pr}(m|y^*)$ is then estimated by
\begin{equation*}
\widehat{\mathrm{Pr}}_\varepsilon(m | y^*) := \frac{1}{T} \sum_{t=1}^{T} \mathbb{I}(m_t = m), \quad m = 1,2,\\
\end{equation*}
where $\mathbb{I}(m_t = m) = 1$ if $m_t = m$ and $\mathbb{I}(m_t = m) = 0$ otherwise. 

For the experiments in our paper, we employed the corresponding $k$-NN (nearest neighbors) version, as recommended by \citet{grelaud2009abc}.

\paragraph{ABC Sequential Monte Carlo (ABC-SMC)}
The method proposed by \citet{toni2009approximate}, called {\em ABC Sequential Monte Carlo (ABC-SMC)}, is also a rejection-based ABC approach to Bayesian model selection, as for ABC-MC.
Based on sequential Monte Carlo, the method iteratively updates the threshold $\varepsilon$ and the associated proposal distributions in an adaptive manner.

\paragraph{ABC Random Forests (ABC-RF)}
The approach by \citet{pudlo2015reliable}, termed {\em ABC Random Forests (ABC-RF)}, is based on  interpretation of Bayesian model selection as a classification problem.
As for ABC-MC, consider the following sampling procedure:
For $t = 1,\dots, n \in \mathbb{N}$, sample model index $m_t \sim \mathrm{Pr}(m)$ from the prior, sample model parameters $\theta_t^m \sim \pi_m(\theta^m)$, and then sample pseudo-data $y_t \sim P_m(y|\theta^m)$.
The posterior probability $\mathrm{Pr}(m|y^*)$ is then the conditional probability of model index $m$ conditioning on observed data $y^*$ that is induced from the joint distribution of $(m_t,y_t)_{t=1}^n$.
Thus, the problem of model selection can be cast as estimation of this conditional probability, or more specifically as classification of observed data $y^*$ to one of the model indices $m = 1,2$ (or class labels) using $(m_t,y_t)_{t=1}^n$ as training data.
ABC-RF solves this classification problem using random forests, thereby providing a way of Bayesian model selection.
The essentially same idea can be used for parameter estimation in the setting of ABC, as proposed by \citet{raynal2016abc}.


\section{Detailed explanation of experimental settings}

\subsection{Hyper-parameter selection}
For each method, we determined its hyper-parameters by adopting a cross-validation-like approach described in \citet[Sec.~4]{Park2015}, unless otherwise stated.
That is, to evaluate one configuration of hyper-parameters, we first used 80\% of the observed data for point estimation and then computed the discrepancy between the rest of the observed data and the ones simulated from point estimates; after applying this procedure to all candidate configurations, the one with the lowest discrepancy was finally selected.

For ABC-MC, the number $k$ of $k$-nearest neighbours was  selected from $\{100,200,300,500,800,1000\}$ using the cross-validation like approach. 
For ABC-SMC, we used the median of pairwise distances between the observed and simulated data as an initial tolerance level. 
As a perturbation kernel, we used the zero-mean Gaussian distribution with variance $0.1$, unless otherwise stated.

For ABC-RF, we set the number of trees to be 500. 
We set the number of bootstrapping samples to be the number of the sample points. 
We set the size of the subset of parameters to be considered at each node in a tree to be $\sqrt{d}$ for classification and $d/3$ for regression, where $d$ is the number of parameters. 
These configurations are those recommended by \citet{pudlo2015reliable}.

For our method we defined a Gaussian kernel $\exp(- \| x - x' \|^2 / \gamma^2)$ for each of observations, parameters and mixing coefficients. 
We selected the bandwidths of these kernels, denoted for now by $\gamma_\mathcal{Y}$, $\gamma_\Theta$ and $\gamma_\phi$, 
 as well as the regularization constant $\delta$ by the above cross-validation like approach.
More specifically, for a constant $s > 0$ let
$\gamma_\mathcal{Y} = s \times \mathrm{med}( y_{N,1},\dots,y_{N,n})$, 
$\gamma_\Theta = s \times \mathrm{med}( (\theta^m_{N,1})_{m=1}^M,\dots, (\theta^m_{N,n})_{m=1}^M)$, 
and $\gamma_\phi = s \times \mathrm{med}( (\phi^m_{N,1})_{m=1}^M, \dots, (\phi^m_{N,n})_{m=1}^M)$, 
where $\mathrm{med} (x_1,\dots,x_n)$ denotes the median of pairwise distances between points $x_1,\dots,x_n$.

We selected this scaling constant $s$ and the regularization constant $\delta$ by the cross validation-like approach, where candidates for $s$ were logarithmically equally-spaced values between $2^{-2}$ and $2^{2}$ and those for $\sigma$ were logarithmically equally-spaced values between $2^{-4}$ and $2^{2}$.



\subsection{Parameter estimation}
For the purpose of comparison, we also performed parameter estimation with the competing methods as follows. 
We first chose a model $m \in \{ 1, \dots, M \}$ based on a model selection method, and then only retained the generated pairs $(y_i^m,\theta_i^m)$ of simulated data and parameters associated with the model $m$.
For ABC-RF, we then performed regression by Random Forests from data to parameters based on the retained pairs $(y_i^m,\theta_i^m)$.
For ABC-MC and ABC-SMC, we estimated the posterior of the parameters based on the retained pairs $(y_i^m,\theta_i^m)$, and then applied the mean shift algorithm \citep{fukunaga1975estimation} to obtain MAP estimates for the parameters; 
for mean-shift, we used the implementation in the scikit-learn, with the bandwidth of KDE being specified as the median of all the pairwise distances of all samples.


\section{Additional experimental results}

\subsection{Mixing coefficients and the effects of the Dirichlet concentration parameter}
\label{sec:poly-qualitative}

We performed simple experiments to see how the mixing coefficients for candidate models in the proposed method behaves. 
To this end, we deal with model selection for a regression problem, where the 3rd order and 10th order polynomials are to be compared. 

Let $x_1,\dots,x_{20} \subset [-1,2]$ be equally-spaced points such that $ -1= x_1 < x_2 < \cdots < x_{20} = 2$.
The 3rd and 10th order polynomial models are then defined as $y_i = \sum_{\ell=0}^3 a_\ell x_i^{\ell} + \varepsilon_i $ and $y_i = \sum_{\ell=0}^{10} b_\ell x_i^{\ell} + \varepsilon_i$, respectively $(i = 1,\dots,20)$, where $(a_\ell)_{\ell=0}^{3} \subset \mathbb{R}$ and $(b_\ell)_{\ell=0}^{10} \subset \mathbb{R}$ are the parameters in the models.
For each parameter in $(a_\ell)_{\ell=0}^{3}$ and $(b_\ell)_{\ell=0}^{10}$ we defined a prior as the uniform distribution over $[-100,100]$. 
When generating observed data $y^* := (y^*_1,\dots,y^*_{20}) \in \mathbb{R}^{20}$, we used either $a_0 = \cdots = a_3 = 4$ (when the 3rd order is true) or
$b_0 = \cdots = b_{10} = 4$ (when the 10th order is true).
We set the concentration parameter of the Dirichlet prior for $\phi_1$ and $\phi_2$ to be $100$. 

\begin{figure}[t]
\begin{center}
\includegraphics[width=\linewidth]{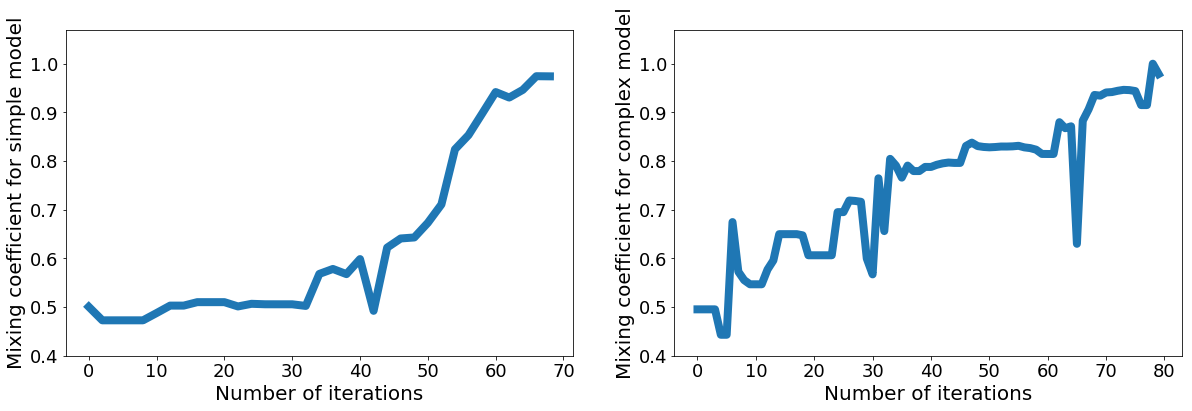}
\end{center}
    \vspace{-3mm}
    \caption{Demonstrations of how the mixing coefficient for the true model changes, as the iteration proceeds. 
    Left: the ground-truth was the 3rd order model (simple); Right: the ground-truth was the 10th order model (complex). (Sec.~\ref{sec:poly-qualitative})}
    \label{fig:iteration_100}
    \vspace{-5mm}
 \end{figure}

Figure \ref{fig:iteration_100} describes how the mixing coefficient for the ground-truth model changed as iteration proceeded, for both cases where the ground truth was the 3rd order (left) or the 10th order (right). 
The results show that for both cases the coefficients gradually converged towards the true model, suggesting the validity of our mixture approach. 

\begin{figure}[t]
 \begin{center}
 \includegraphics[width=\linewidth]{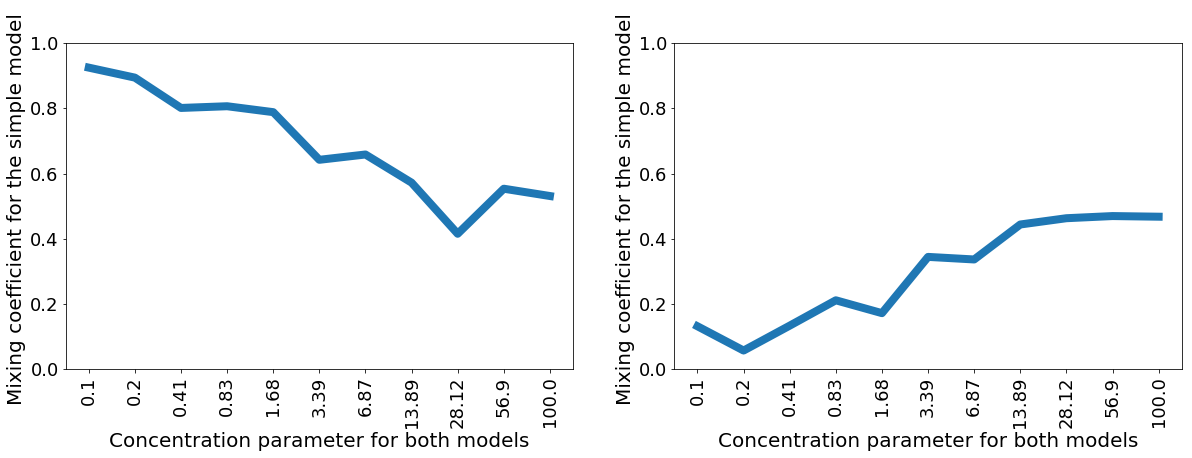}
 \end{center}
 \vspace{-3mm}
    \caption{The coefficients for the simpler model (i.e., the 3rd order polynomial) after 30 iterations, for different values of the concentration parameter $\alpha$. 
    Left: the truth was the 3rd order model; 
    Right: the truth was the 10th polynomial model. (Sec.~\ref{sec:poly-qualitative})}
    \label{fig:iteration_fixed}
    \vspace{-5mm}
 \end{figure}

We next investigated how the behavior of the proposed method changes when changing the concentration parameter of the Dirichlet prior $\alpha$. 
Figure \ref{fig:iteration_fixed} shows the mixing coefficients after $30$ iterations, when running our method for different values of $\alpha$. 
The results show that for smaller $\alpha$, the coefficients tend to be more confident about the true model (i.e., the coefficients tend to be close to $0$ or $1$).
For instance, in Figure \ref{fig:iteration_fixed} (a) on the case where the truth is the 3rd order, the coefficient for the 3rd order polynomial tends to $1$ as $\alpha$ decreases (i.e., more confident the truth is the 10th-order), and approaches $0.5$ as $\alpha$ increases (more neutral).
This is consistent with the role of the concentration parameter for the Dirichlet prior: If $\alpha$ is small, then the mixing coefficients sampled from the Dirichlet prior tend to take values close to either of $0$ or $1$, while if $\alpha$ is large, then the sampled mixing coefficients tend to be close to 0.5.

\begin{figure}[t]
 \begin{center}
 \includegraphics[width=\linewidth]{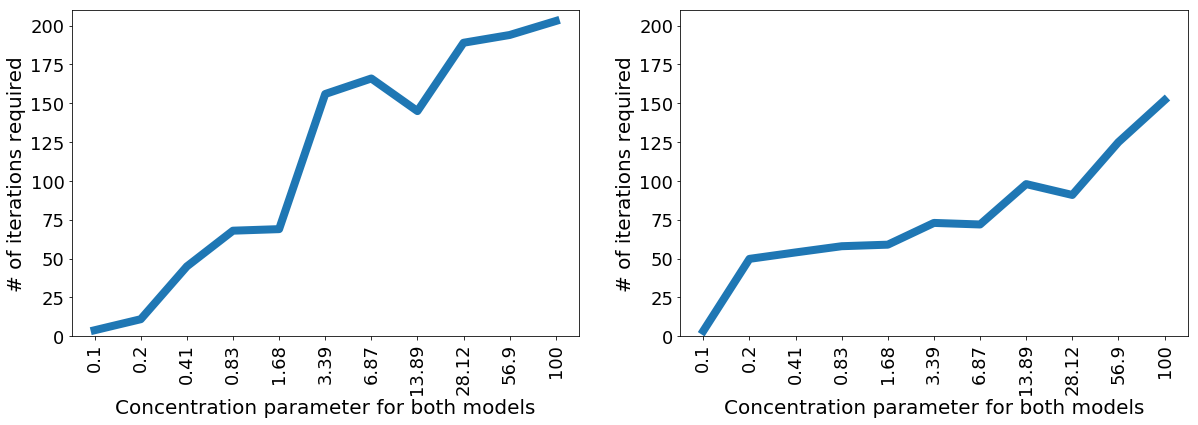}
 \end{center}
 \vspace{-3mm}
    \caption{The number of recursive Bayes' updates required to make the mixing coefficient larger 0.95 for a true model over a varying concentration parameter $\alpha$.
    Left: the truth is the 3rd order model; 
    Right: the truth is the 10th order model. (Sec.~\ref{sec:poly-qualitative})}
    \label{fig:num_convergence}
    \vspace{-5mm}
 \end{figure}

We nevertheless found that, regardless of the choice of the concentration parameter, the mixing coefficients converge to either $0$ or $1$, as the iteration proceeds. 
This is described in Figure \ref{fig:num_convergence}, which plots the number of iterations required for the mixing coefficients to reach the region $[0,0.05]$ or $[0.95,1]$.
The result shows that a larger concentration parameter leads to rather conservative decisions regarding model selection, requiring a larger number of iterations for convergence. 



\subsection{Extrapolation errors in polynomial experiments (Sec.~4.2)}

To evaluate the generalization performances of the parameters (and the models) estimated by each method, we also report extrapolation errors given as follows. 
Let $x_{26},\dots,x_{30} \subset [5,6]$ be equally-spaced points such that $5 = x_{26}  < \cdots < x_{30} = 6$.
We then generated observed data $y^* := (y^*_{26},\dots,y^*_{30}) \in \mathbb{R}^{5}$ from the ground-truth model evaluated at $x_{26},\dots,x_{30}$. 
We calculated the corresponding data errors at $x_{26},\dots,x_{30}$ for each method, in the same way as the one explained in the main body.

\begin{figure}[htbp] 
\begin{center}
\includegraphics[width=0.6\linewidth]{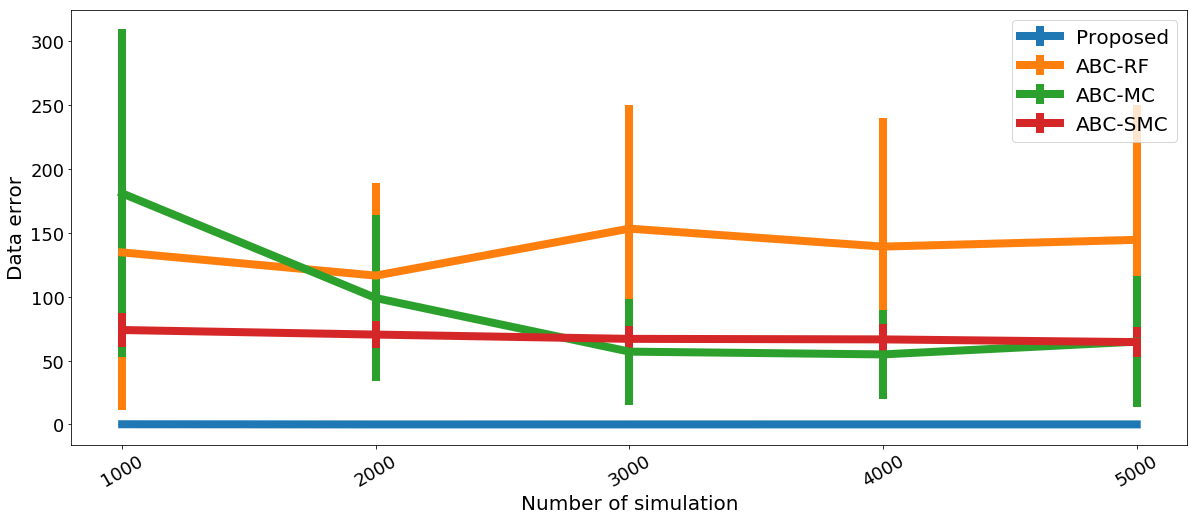}
\includegraphics[width=0.6\linewidth]{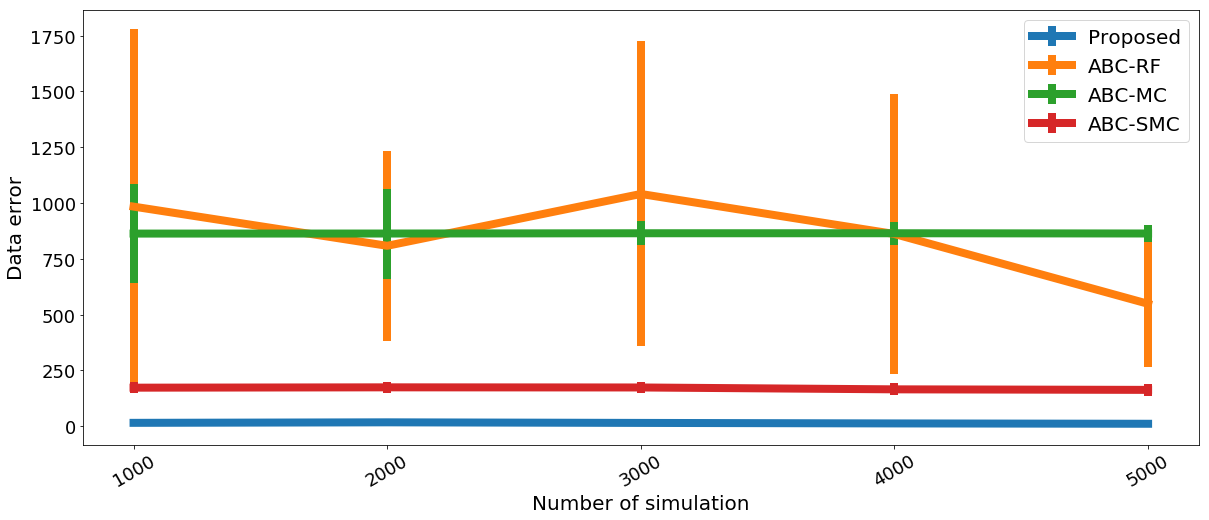}
\caption{Extrapolation errors for polynomial experiments with appropriate prior distributions. Top: the ground truth is the 3th order polynomial model; Bottom: the ground truth is the 4th order polynomial model. }
\label{fig:poly_gen_appropriate}
\end{center}
 \end{figure}

\begin{figure}[htbp] 
\begin{center}
\includegraphics[width=0.60\linewidth]{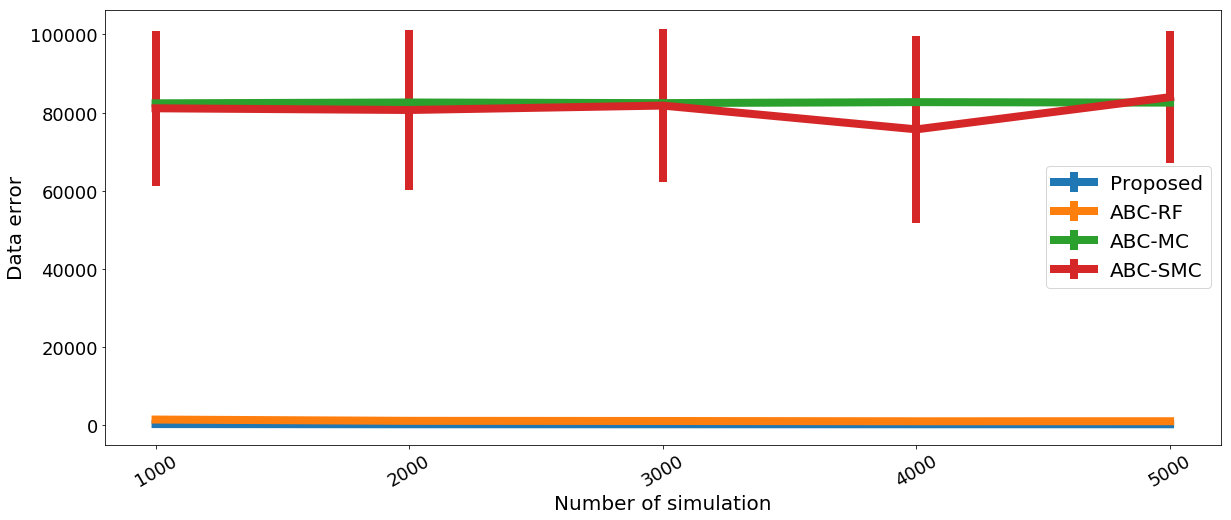}
\includegraphics[width=0.60\linewidth]{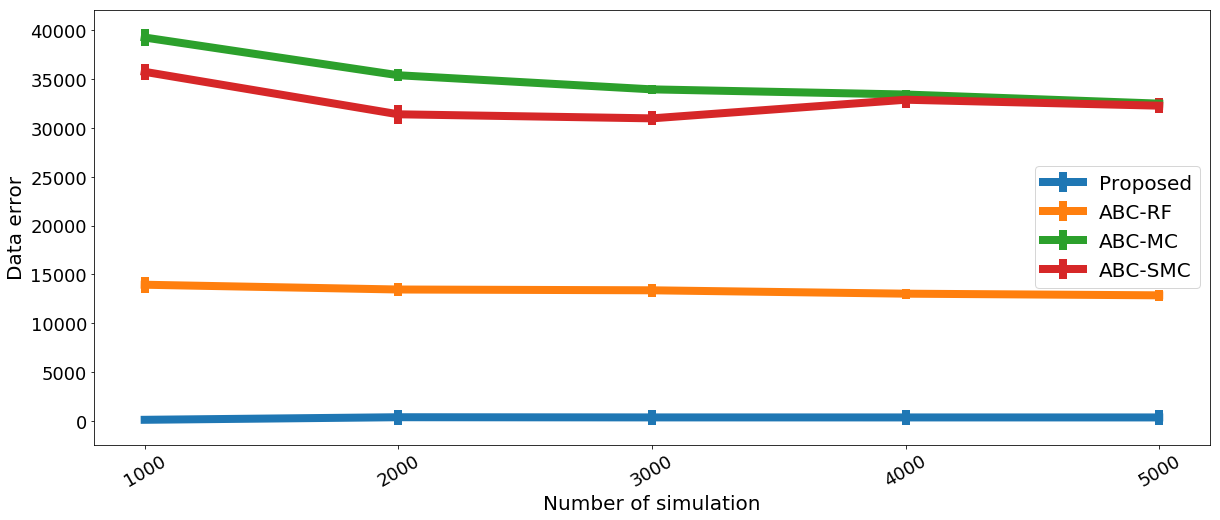}
\caption{Extrapolation errors for polynomial experiments with misspecified prior distributions. Top: the ground truth is the 3th order polynomial model; Bottom: the ground truth is the 4th order polynomial model. }
\label{fig:poly_gen_misspecifed}
\end{center}
 \end{figure}

We show the results for the appropriate prior distributions in Figure \ref{fig:poly_gen_appropriate} and for the misspecified prior distributions in Figure \ref{fig:poly_gen_misspecifed}.
The proposed method outperformed the other methods in all the experiments. 
Notably, in the misspecified case, the errors of the other methods are huge, while those of the proposed method are negligible, verifying that the proposed method successfully selected the true model and estimated its parameters.


\subsection{Predator-prey models (Sec.~4.3)}
\begin{figure}[htbp]
  \begin{center}
\includegraphics[width=0.45\linewidth]{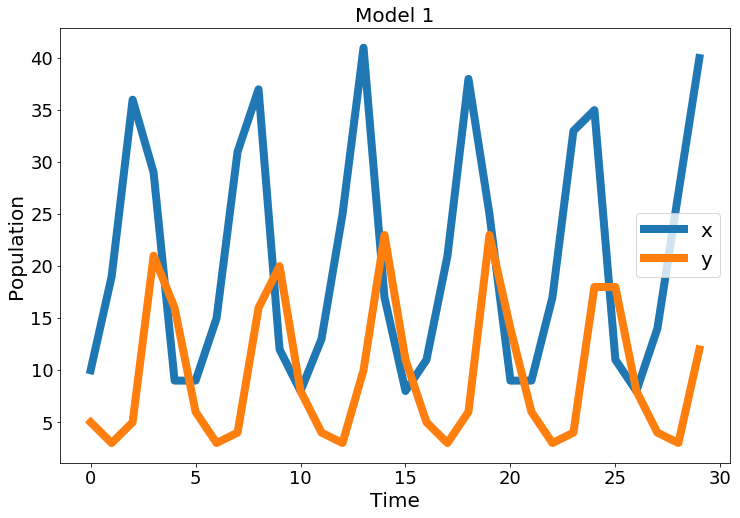}
\includegraphics[width=0.45\linewidth]{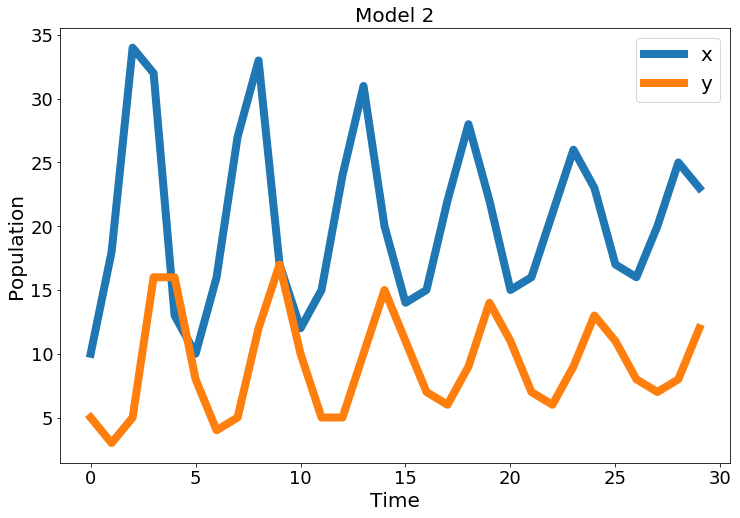}
  \end{center}
    \caption{Observed trajectories (with noises) of the two populations (predators $y(t)$ and prey $x(t)$) generated with the true parameters.
    Left: Model 1; Right: Model 2.}
    \label{fig:lr_1_dynamics}
 \end{figure}

\paragraph{Dynamics of the two models.}
We show observed trajectories of the two models (with noises added) in Figure \ref{fig:lr_1_dynamics}.  

\paragraph{Experiments in a more difficult setting.}
To make the problem more challenging, we assumed the initial population sizes $x(0)$ and $y(0)$ to be unknown, and let them as parameters in the models to be estimated. We defined a wide prior distribution for each of $x(0)$ and $y(0)$ as the uniform distribution over $[0,2000]$. 
We set $T = 30$, and removed the first 10 values $x(0),...,x(9)$ and $y(0),...,y(9)$ from observed data. 
In preliminary experiments we found that, because of the wide prior distribution for the initial values, the variance of the generated data could become so huge that any of the methods did not work well.
We therefore applied the arctangent transformation to the observed and simulated data to stabilize the large variance, following \citet[p.854]{yildirim2015parameter}.

\begin{figure}[t]
  \begin{center}
\includegraphics[width=0.8\linewidth]{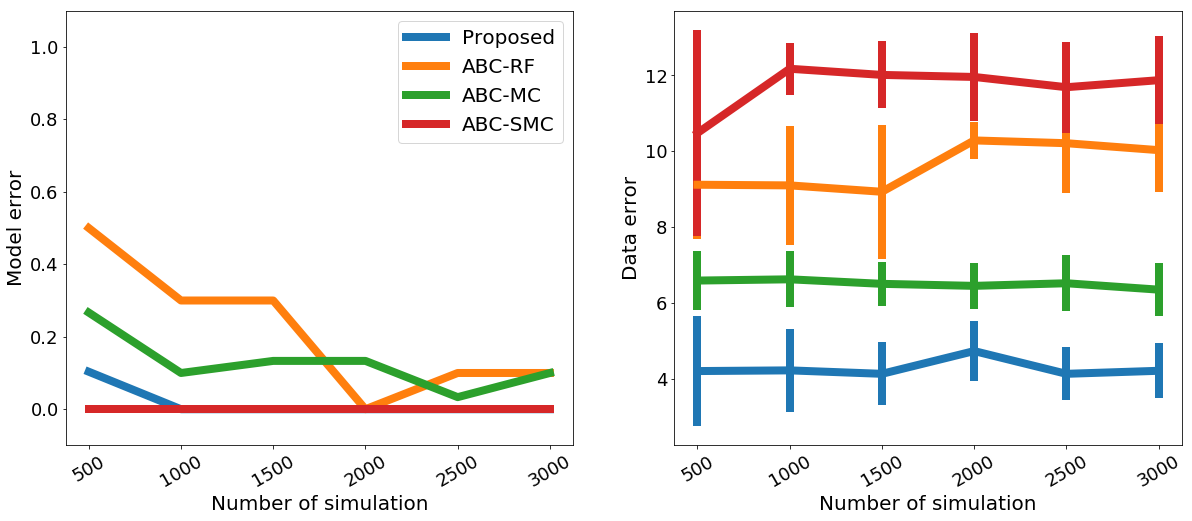}
  \end{center}
    \caption{Model errors (left) and data errors (right) in the predator-prey experiments on the difficult setting.
    The ground-truth is Model 1.}
      \label{fig:predator-prey-simpler-truth-wide}
 \end{figure}

\begin{figure}[t]
  \begin{center}
\includegraphics[width=0.8\linewidth]{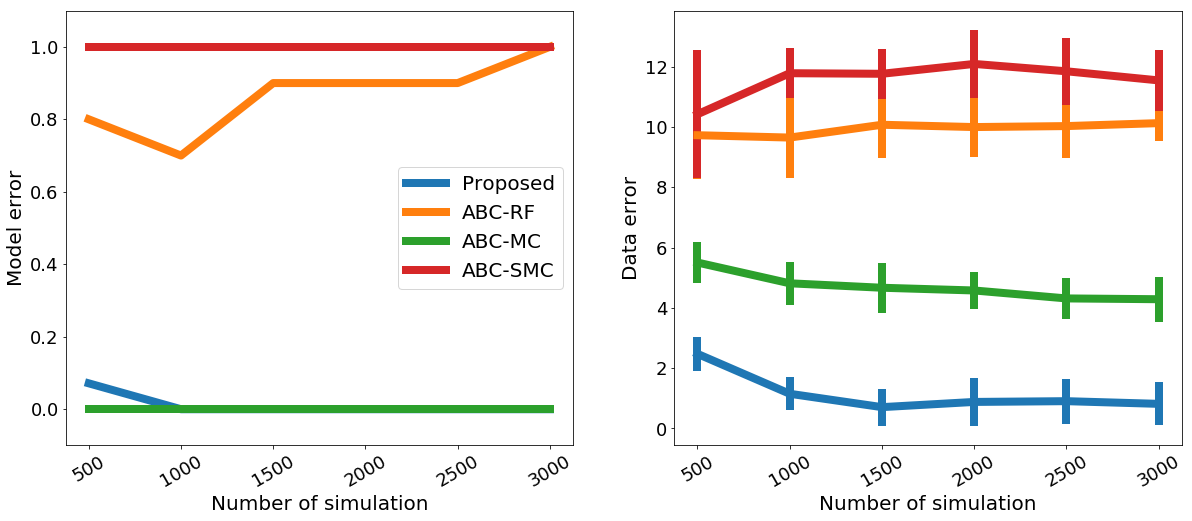}
  \end{center}
    \caption{Model errors (left) and data errors (right) in the predator-prey experiments on the difficult setting.
    The ground-truth is Model 2.}
      \label{fig:predator-prey-complex-truth-wide}
 \end{figure}

Figures \ref{fig:predator-prey-simpler-truth-wide} and \ref{fig:predator-prey-complex-truth-wide} respectively show the results for the two cases where the ground truth is Model 1 or Model 2. 
ABC-SMC and ABC-MC estimated the true model correctly only in one of the two cases. 
In other words, those methods favor one model exclusively. 
Only the proposed method was able to estimate the true model reliably for both cases.
The proposed method also outperformed other methods in terms of data errors for both cases.


\begin{figure}[htbp]
  \begin{center}
\includegraphics[width=0.6\linewidth]{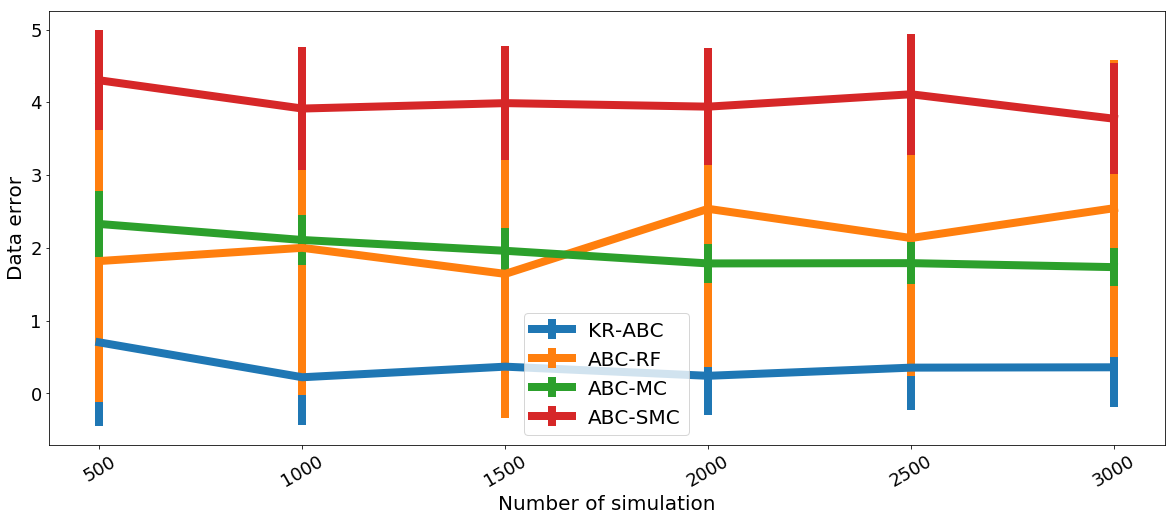}
\includegraphics[width=0.6\linewidth]{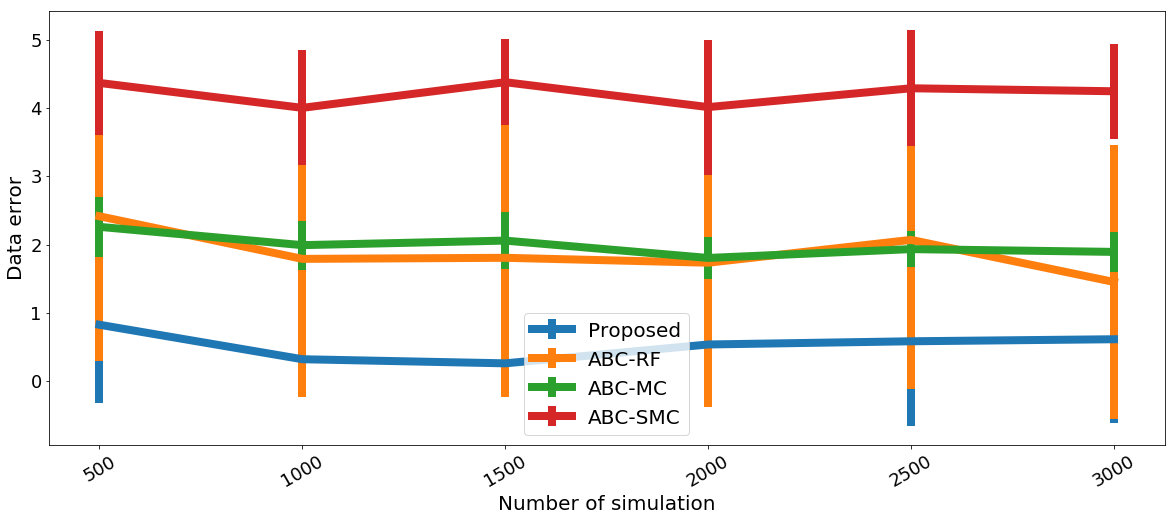}
  \end{center}
    \caption{Extrapolation errors in the predator-prey experiments for the setting of the main body.
    Top: the ground-truth is Model 1; Bottom: the ground-truth is Model 2.}
    \label{fig:lr_1_gen}
 \end{figure}

\begin{figure}[htbp]
  \begin{center}
\includegraphics[width=0.6\linewidth]{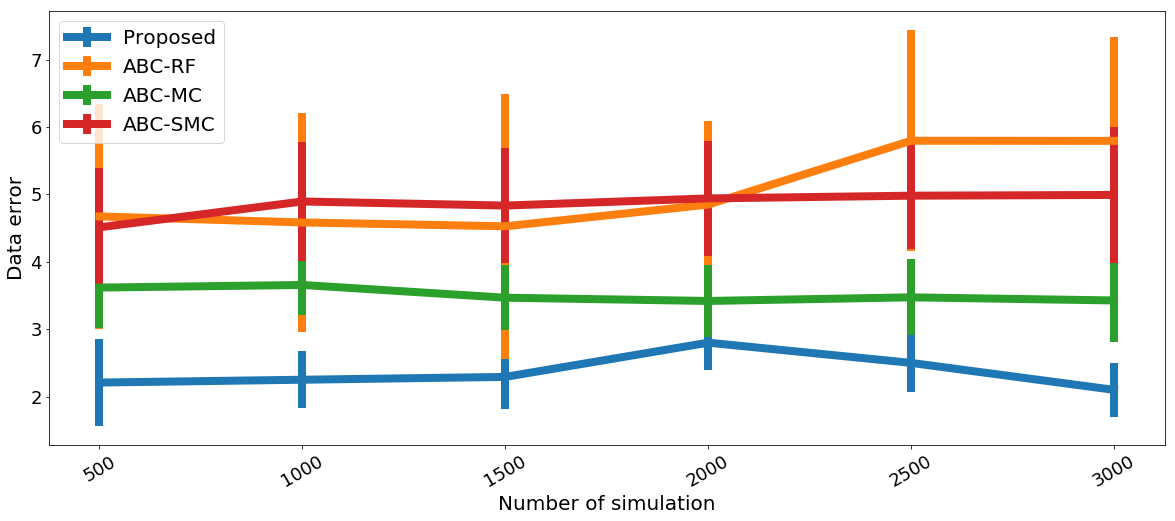}
\includegraphics[width=0.6\linewidth]{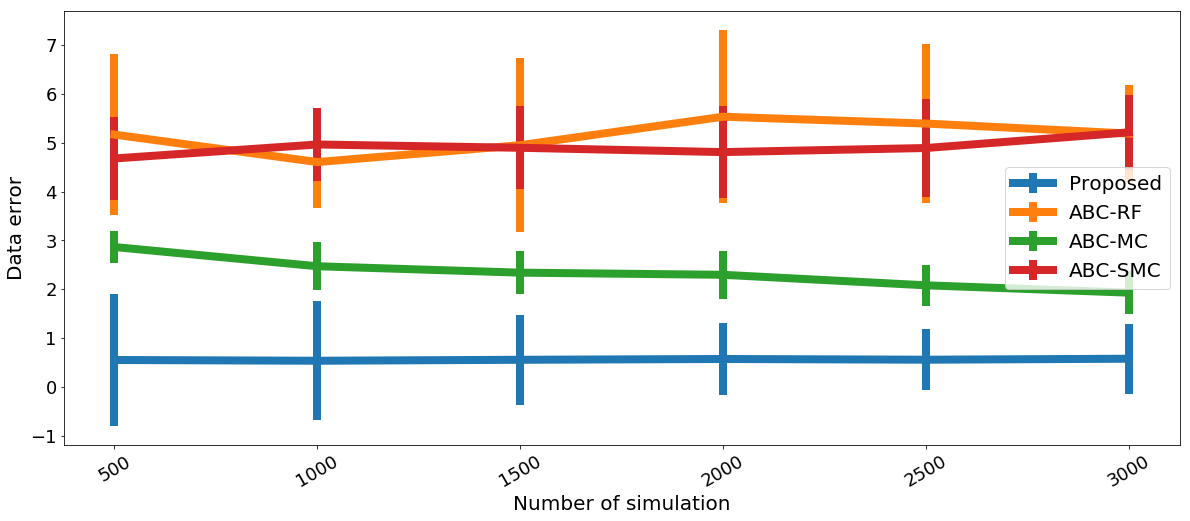}
  \end{center}
    \caption{Extrapolation errors in the predator-prey experiments for the difficult setting.
    Top: the ground-truth is Model 1; Bottom: the ground-truth is Model 2.}
    \label{fig:lr_1_wide_gen}
 \end{figure}

\paragraph{Extrapolation errors}
To evaluate the predictive performances of the parameters (and the models) estimated by each method, we also report extrapolation errors, given as follows.
Extrapolation errors were evaluated at equally-spaced 5 points $t = 21,22,\dots,25$ for the setting of the main body, and $t = 31,32,\dots, 35$ for the difficult setting. 
Results for the setting of the main body are shown in Figure \ref{fig:lr_1_gen}, and those for the difficult setting are shown in Figure \ref{fig:lr_1_wide_gen}.
The proposed method outperformed the other methods in all the experiments.

\subsection{Dynamical systems for epidemics (Sec.~4.4)} 

\begin{figure}[htbp]
  \begin{center}
\includegraphics[width=0.5\linewidth]{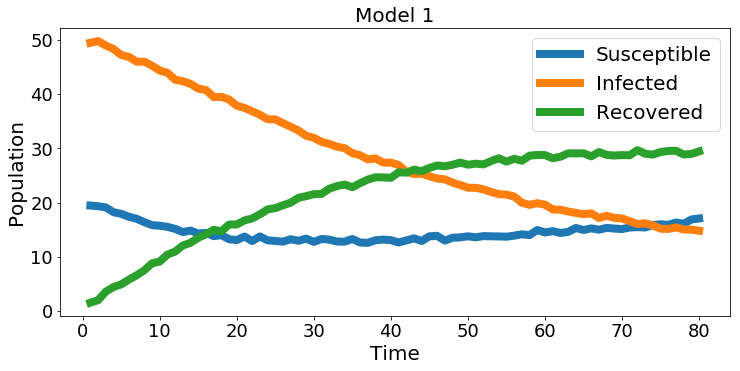}
\includegraphics[width=0.5\linewidth]{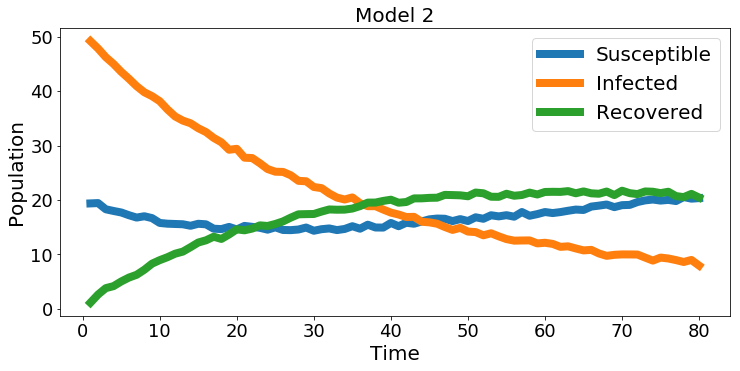}
\includegraphics[width=0.5\linewidth]{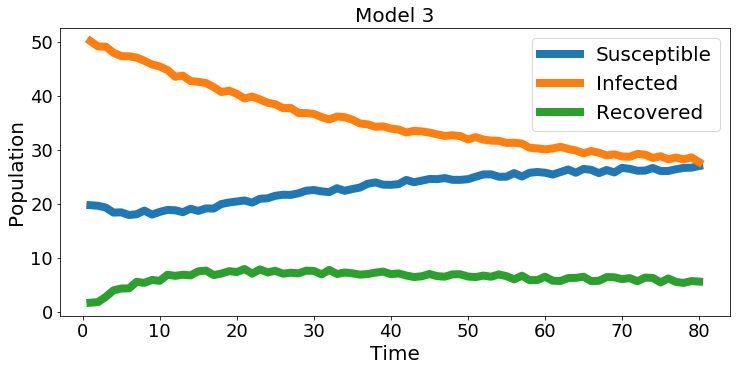}
  \end{center}
    \caption{Observed trajectories from Model 1 (Top), Model 2 (Middle) and Model 3 (Bottom) with the true parameters in epidemics experiments. 
    As explained in the main body, independent noises are added in each trajectory. }
    \label{fig:sir_trajectories}
 \end{figure}

\paragraph{Dynamics of the three models.}
We show observed trajectories of the three models (with noises added) in Figure \ref{fig:sir_trajectories}.

\begin{figure}[htbp]
  \begin{center}
\includegraphics[width=0.8\linewidth]{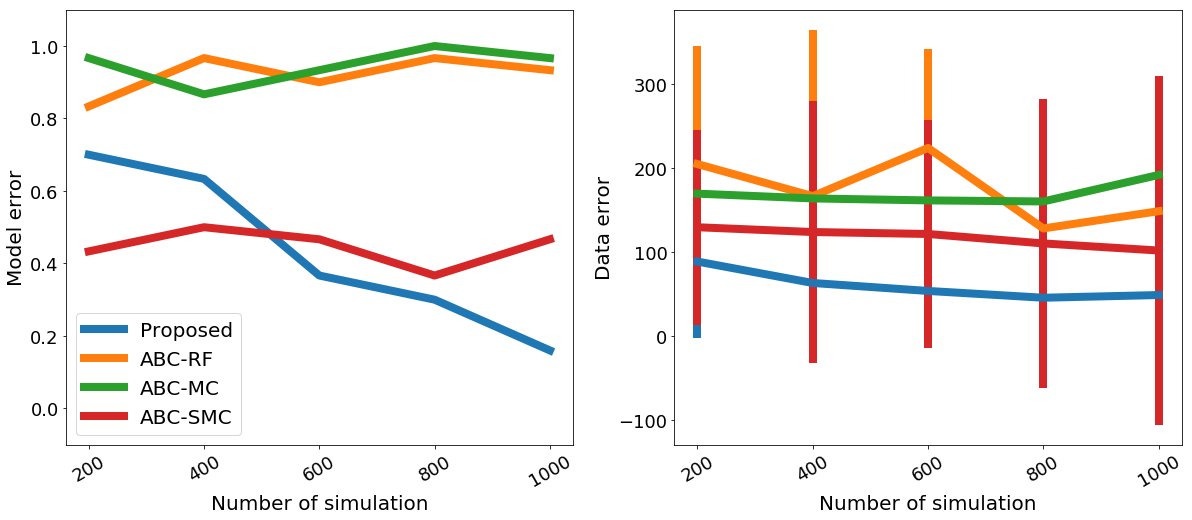}
  \end{center}
    \caption{Model errors (left) and data errors (right) in the epidemics experiments for the difficult setting. The ground-truth model is Model 2.}
    \label{fig:sir_2_wide}
 \end{figure}

\begin{figure}[htbp]
  \begin{center}
\includegraphics[width=0.8\linewidth]{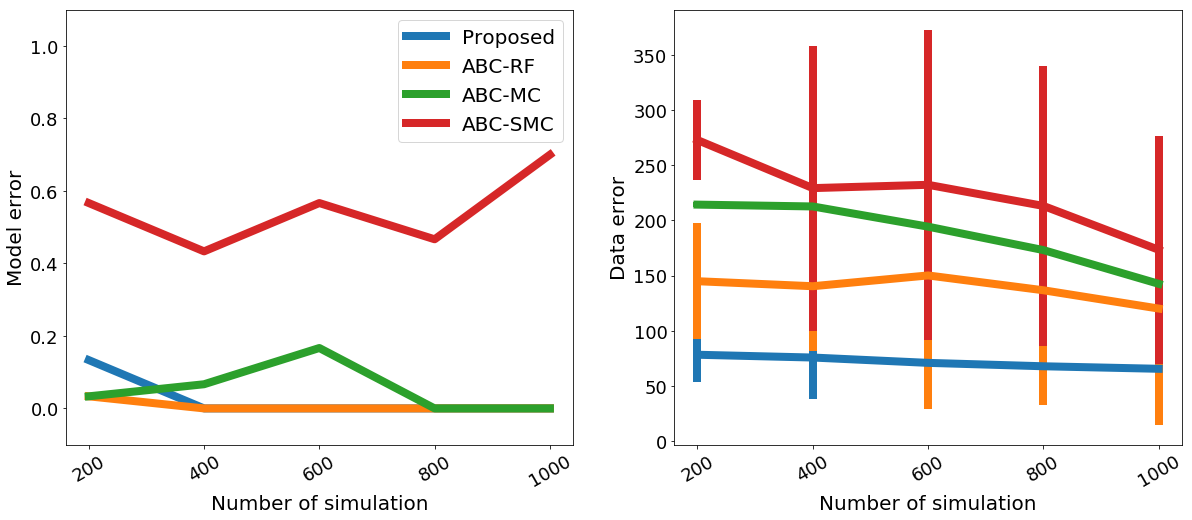}
  \end{center}
    \caption{Model errors (left) and data errors (right) in the epidemics experiments for the difficult setting. The ground-truth model is Model 3.}
    \label{fig:sir_3_wide}
 \end{figure}

\paragraph{Experiments in a more difficult setting.}
To make the problem more challenging, we treated the initial values $S(0)$, $I(0)$ and $R(0)$ (and $L(0)$ for Model 2) as unknown parameters in the models to be estimated.
We set $T = 80$, and removed the first 10 values $S(0),...,S(9), I(0),..., I(9), R(0),...,R(9)$ from observed data. 
We defined a wide prior for each of $S(0)$, $I(0)$, $R(0)$ and $L(0)$ as the uniform distribution over $[0,500]$. 
As none of the methods was able to identify a true model if the three candidate models were jointly compared, we report only the case where the second and third models are compared. 
The results are shown in Figure \ref{fig:sir_2_wide} and \ref{fig:sir_3_wide}. 
ABC-RF and ABC-MC identified correctly the ground-truth model only for one of the two cases, and failed for the other case. Only the proposed method was able to identify the true model reliably for the both cases.


\begin{figure}[htbp]
  \begin{center}
\includegraphics[width=0.6\linewidth]{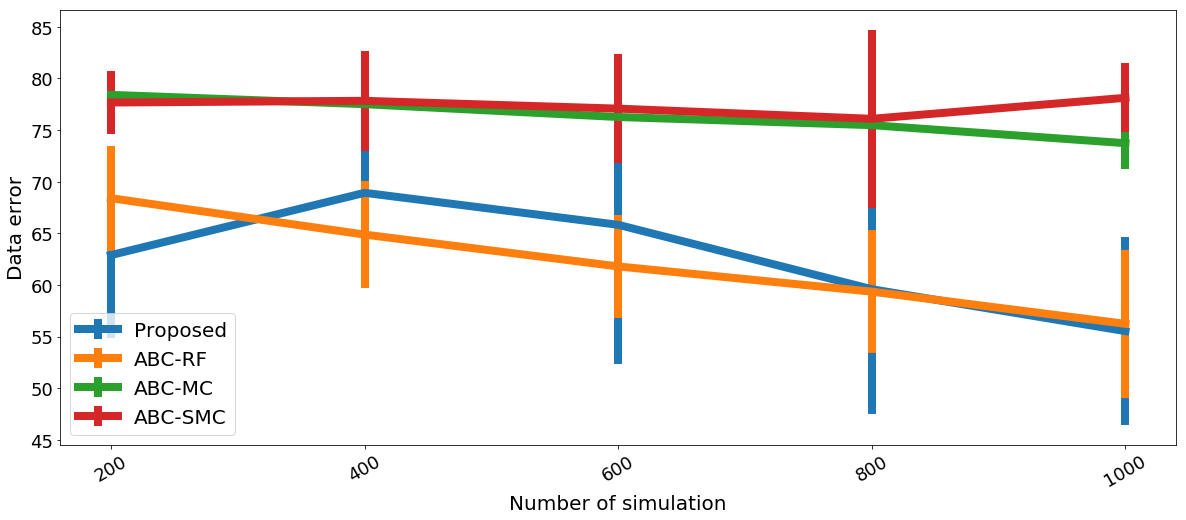}
\includegraphics[width=0.6\linewidth]{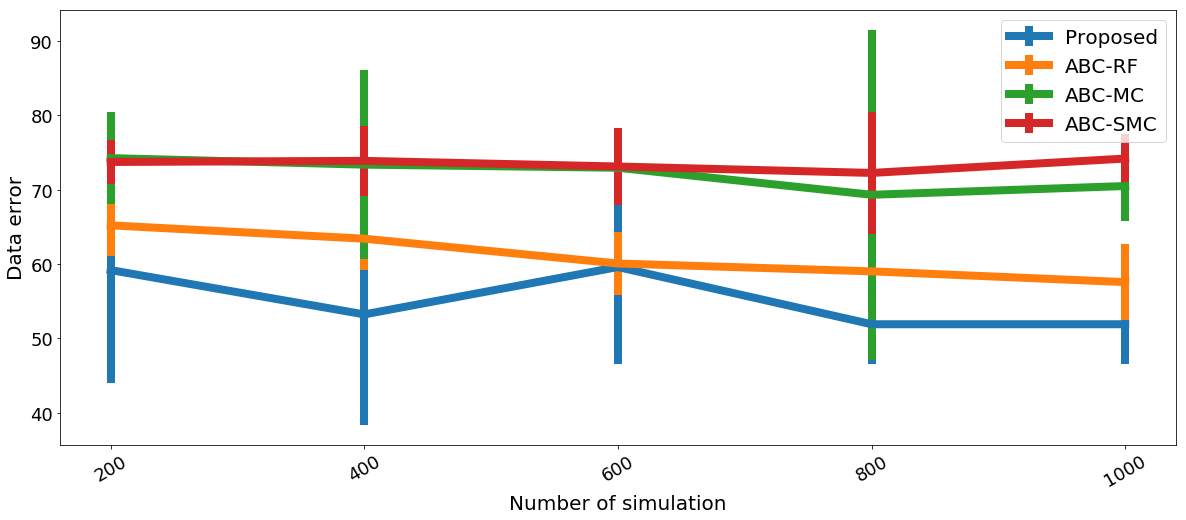}
\includegraphics[width=0.6\linewidth]{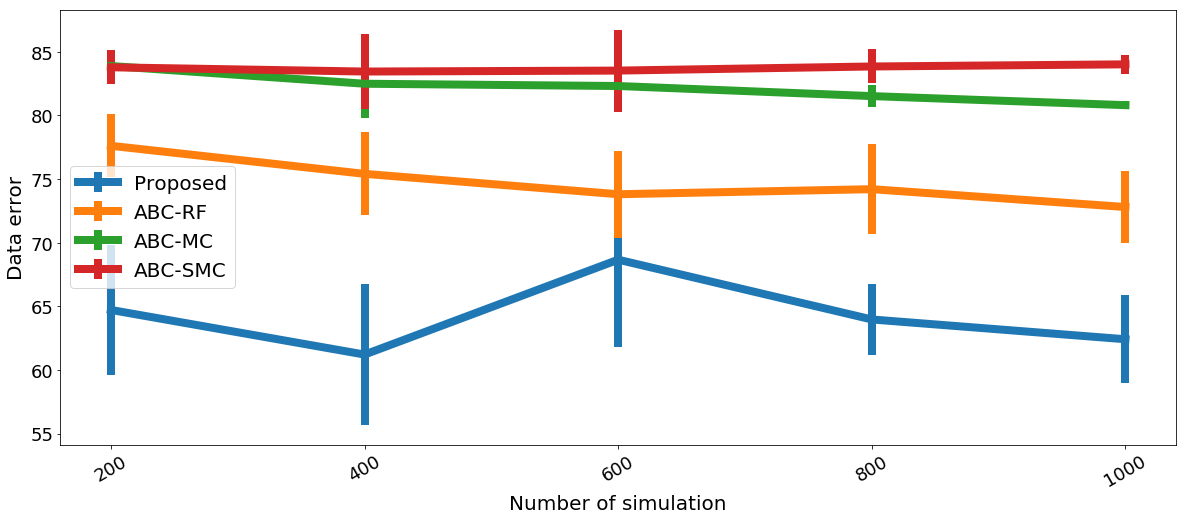}
  \end{center}
    \caption{Extrapolation (or prediction) errors in the epidemics experiments, for the cases where the ground-truth is Model 1 (top), Model 2 (middle) and Model 3 (bottom). 
    The settings (such as the prior distributions) are the same as in the main body. }
    \label{fig:sir_1_gen}
 \end{figure}

\begin{figure}[htbp]
  \begin{center}
\includegraphics[width=0.6\linewidth]{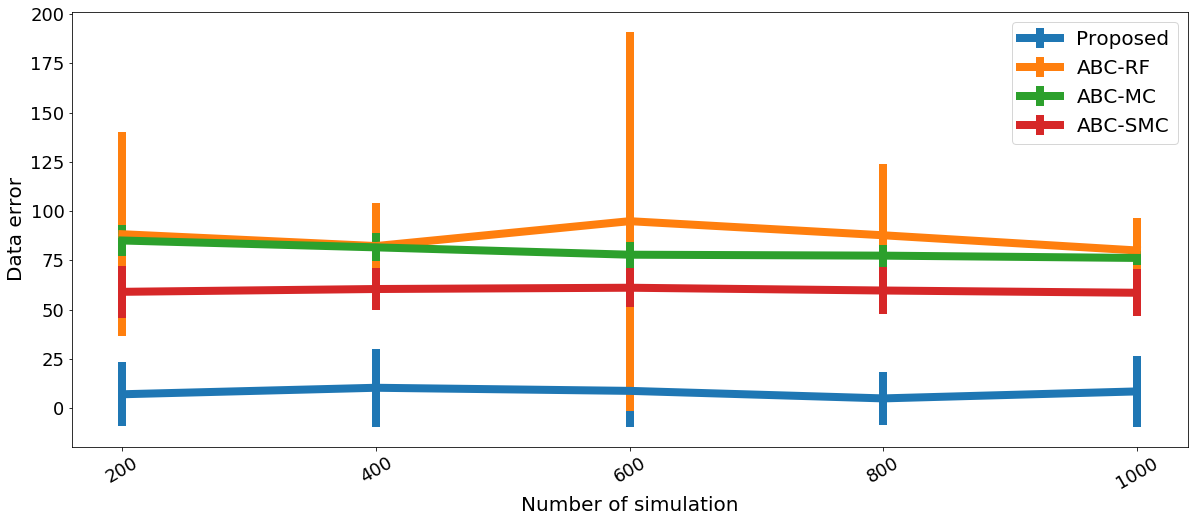}
\includegraphics[width=0.6\linewidth]{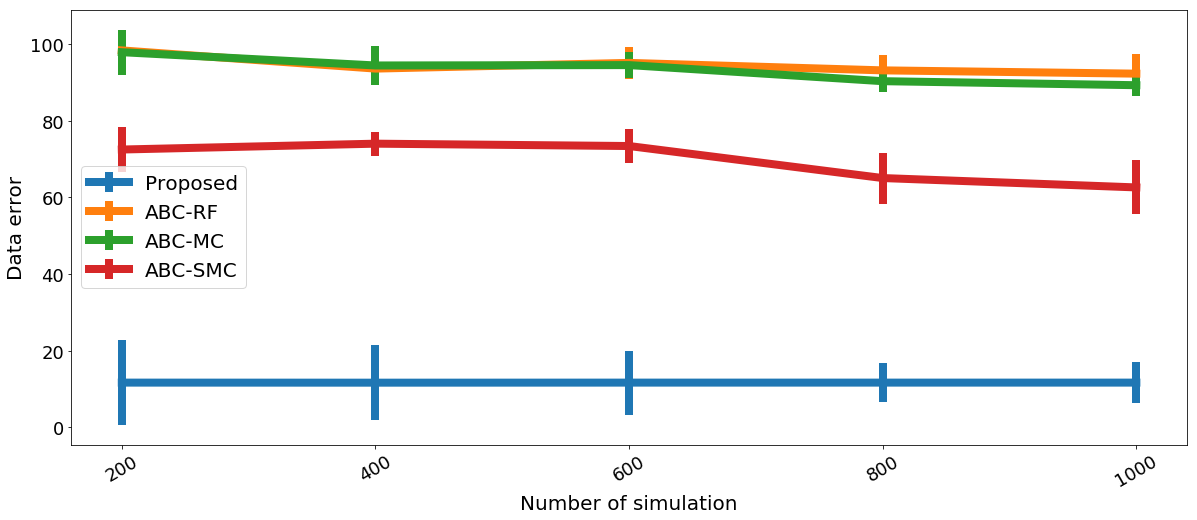}
  \end{center}
    \caption{Extrapolation (or prediction) errors in the epidemics experiments, for the cases where the ground-truth is  Model 2 (top) and Model 3 (bottom). 
    The setting is the difficult setting defined in the Supplementary Materials.}
    \label{fig:sir_2_wide_gen}
 \end{figure}

\paragraph{Extrapolation errors.}
The extrapolation errors were evaluated at equally-spaced 15 points $t = 71, 72, \dots, 85$ for the setting of the main body, and at $t = 81, 82, \dots, 95$ for the difficult setting defined above.
Results for the setting of the main body are shown in Figure \ref{fig:sir_1_gen}, and those for the difficult setting are shown in Figure \ref{fig:sir_2_wide_gen}.
The proposed method outperformed the other methods for most of the experiments.

\end{document}